\documentclass[10pt,twocolumn,letterpaper]{article}
\usepackage{cvpr}
\usepackage[T1]{fontenc}
\usepackage{tabularx,array}
\usepackage{tikz}
\usetikzlibrary{arrows.meta,positioning,calc}
\usepackage{microtype}

\ExplSyntaxOn

\clist_new:N \l_bench_first_clist
\clist_new:N \l_bench_last_clist
\int_new:N \l_bench_max_int
\int_new:N \l_bench_pos_int

\cs_new_protected:Npn
  \bench_gradient_engine:nnnnnnnn #1#2#3#4#5#6#7#8
  {
    \int_set:Nn \l_bench_max_int { \str_count:n {#1} }
    \int_zero:N \l_bench_pos_int

    \int_step_inline:nnn {1} { \l_bench_max_int }
      {
        \exp_args:Ne
        \str_if_eq:nnTF { \str_item:nn {#1} {##1} } {~}
          { }
          { \int_incr:N \l_bench_pos_int }

        \color_select:nn {#8}
          {
            \int_eval:n
              {
                (
                  \l_bench_pos_int * #5
                  + (\l_bench_max_int - ##1) * #2
                ) / \l_bench_max_int
              },
            \int_eval:n
              {
                (
                  \l_bench_pos_int * #6
                  + (\l_bench_max_int - ##1) * #3
                ) / \l_bench_max_int
              },
            \int_eval:n
              {
                (
                  \l_bench_pos_int * #7
                  + (\l_bench_max_int - ##1) * #4
                ) / \l_bench_max_int
              }
          }

        \str_item:nn {#1} {##1}
      }
  }

\NewDocumentCommand \benchgradient {mmmm}
  {
      \group_begin:

    \clist_set:Nn \l_bench_first_clist {#3}
    \clist_set:Nn \l_bench_last_clist {#4}

    \bench_gradient_engine:nnnnnnnn {#2}
      { \clist_item:Nn \l_bench_first_clist {1} }
      { \clist_item:Nn \l_bench_first_clist {2} }
      { \clist_item:Nn \l_bench_first_clist {3} }
      { \clist_item:Nn \l_bench_last_clist {1} }
      { \clist_item:Nn \l_bench_last_clist {2} }
      { \clist_item:Nn \l_bench_last_clist {3} }
      {#1}
      
    \group_end:

  }

\NewDocumentCommand \gradientHSB {mmm}
  {
    \benchgradient {HSB} {#1} {#2} {#3}
  }

\ExplSyntaxOff

\DeclareRobustCommand{\bench}{%
  \gradientHSB{3D GameDev Arena}{0,240,240}{240,240,240}%
}

\definecolor{primary}{RGB}{20,68,132}
\newcommand{\takeaway}[2]{%
  \par\vspace{3.5pt}\noindent
  {\color{primary}\vrule width 2.4pt}%
  \colorbox{primary!8}{%
    \parbox{\dimexpr\linewidth-2.4pt-2\fboxsep\relax}{%
      \vspace{2.2pt}\raggedright\textbf{\color{primary}#1:}\hspace{0.35em}\textbf{#2}\vspace{2.2pt}%
    }%
  }\par\vspace{2pt}\noindent\ignorespaces
}

\newcommand{\yes}{\ensuremath{\checkmark}}
\newcommand{\no}{\ensuremath{\times}}
\newcommand{\nr}{\textemdash}
\newcolumntype{Y}{>{\raggedright\arraybackslash}X}

\makeatletter
\IfFormatAtLeastTF{2025-06-01}{%
  \RemoveFromHook{build/column/before}[lineno]
  \AtBeginDocument{%
    \def\@LN@makecol{%
      \@LN@orig@makecol
      \setbox\@outputbox\vbox{%
        \boxmaxdepth\@maxdepth
        \protected@write\@auxout{}{%
          \string\def\string\@LN@column{\if@firstcolumn1\else2\fi}%
        }%
        \box\@outputbox
      }%
    }%
  }%
}{}
\makeatother

\definecolor{cvprblue}{rgb}{0.21,0.49,0.74}
\usepackage[pagebackref,breaklinks,colorlinks,allcolors=cvprblue]{hyperref}
\usepackage[most]{tcolorbox}
\usepackage{pifont}
\usepackage{enumitem}

\definecolor{promptframe}{RGB}{46, 139, 87}       
\definecolor{promptbg}{RGB}{240, 250, 244}        
\definecolor{promptinnerframe}{RGB}{52, 130, 96}  
\definecolor{promptinnerbg}{RGB}{222, 244, 231}   

\newtcolorbox{promptbox}[2][]{
  enhanced,
  breakable,
  colback=promptbg,
  colframe=promptframe,
  boxrule=1.3pt,
  arc=4.5pt,
  left=10pt, right=10pt, top=12pt, bottom=9pt,
  before skip=14pt, after skip=10pt,
  fonttitle=\bfseries\small,
  coltitle=white,
  title={~#2},
  attach boxed title to top left={xshift=10pt, yshift=-\tcboxedtitleheight/2},
  boxed title style={
    sharp corners,
    colback=promptframe,
    colframe=promptframe,
    boxrule=0pt,
    left=6pt, right=8pt, top=2.5pt, bottom=2.5pt
  },
  #1
}

\newtcolorbox{promptinnerbox}[1][]{
  enhanced,
  colback=promptinnerbg,
  colframe=promptinnerframe,
  boxrule=0.9pt,
  arc=3.5pt,
  left=8pt, right=8pt, top=6pt, bottom=6pt,
  before skip=6pt, after skip=6pt,
  fontupper=\small,
  #1
}
\hypersetup{
  pdftitle={3D GameDev Arena: Benchmarking Coding Agents That Plan, Build, and Playtest extreme Long-Horizon Browser Games},
  pdfsubject={3D GameDev Arena: Benchmarking Coding Agents That Plan, Build, and Playtest extreme Long-Horizon Browser Games},
  pdfauthor={}
}

\def\confName{CVPR}

\def\confYear{2027}
\title{3D GameDev Arena: Benchmarking Coding Agents That Plan, Build, and Playtest extreme Long-Horizon Browser Games}
\author{
Bizhe Bai\textsuperscript{1,2},
Jiliang Jian\textsuperscript{2,3},
Yuqing Yan\textsuperscript{1},
Tao Chen\textsuperscript{1},
Ying Wen\textsuperscript{4}
\\
\textsuperscript{1}Fudan University
\quad
\textsuperscript{2}Shanghai Innovation Institute
\\
\textsuperscript{3}Xi'an Jiaotong University
\quad
\textsuperscript{4}Shanghai Jiao Tong University
}

\begin{document}
\maketitle

\begin{abstract}
Coding agents now resolve repository issues and generate compact playable games, but whether they can build a large 3D game from scratch remains open. Such games tightly couple mechanics, scenes, assets, and interaction logic, and their quality rests on player enjoyment, which existing benchmarks rarely assess. Thus, we introduce \bench{}, a first extreme long-horizon benchmark of 108 from-scratch game-development tasks across 14 genres. 
Given only a text specification and UI/UX reference images, a fully automated plan--implement--test--repair pipeline builds each task into a self-contained browser game. Runs average $\sim$1,000 LLM calls ; the longest uses 14,407 calls, 2.13B  tokens, and 608 assets. Evaluation spans (1) objective code-level functional acceptance against the game specification, (2) agent-based QA and playtesting of 3D presentation and playability, and (3) blind pairwise Bradley-Terry human preference in a public arena. 
The preliminary findings indicate that long-horizon 3D game development remains far from solved, and code-level correctness alone does not dictate player preference.
The public arena is hosted online.\footnote{\url{www.baibizhe.github.io/gamearena/}}

\end{abstract}

\section{Introduction}
\label{sec:intro}
Coding agents have climbed a steep ladder of software engineering tasks: they resolve real GitHub issues in existing repositories \citep{swebench2024,huang2026deepswemeasuringfrontiercoding}, execute complex terminal workflows \citep{terminalbench2026}, and sustain autonomous development of full applications over multiple days \citep{hoh2026}. Building games is the next frontier. Recent benchmarks show that agents can turn a prompt into a compact playable game \citep{gamecraft2026,gamexpert2026,gameasg2026,swegame2026}. But compiling cleanly and running without crashes is only the baseline: the ultimate measure of a game is whether it is fun. This raises a sharper question: \\ \emph{Can coding agents build a large 3D game from scratch, and will the resulting game actually be fun to play?}

Answering this question requires a benchmark that reflects real 3D game development, yet existing benchmarks fall short in three ways (Table~\ref{tab:related}). \textbf{First, real development builds a 3D game from scratch.} An agent should start from an empty folder, design its own codebase, and organize dozens of 3D assets into a working world. Instead, prior benchmarks hand agents ready-made templates \citep{webgamebench2026,cegame2026}, pre-loaded 2D asset packs \citep{gamecraft2026}, simple 3D shapes, built-in test hooks \citep{gameasg2026,swegame2026,cegame2026}, or even a working reference game \citep{swegame2026,gamexpert2026}.
\textbf{Second, real 3D development takes hours and about a thousand tool calls.} Without templates or starter assets, building and debugging a playable 3D game is a long-horizon process. Yet existing benchmarks cap a run at one to two hours or at a few hundred steps (Table~\ref{tab:related}), leaving this scale untested.
\textbf{Third, games should be fun to play.} A game can pass every unit test and still feel boring; whether it is fun is subjective and should be judged by \textbf{human preference}. Yet existing benchmarks rely mainly on code checks, rule checks, and model judges, and rarely measure human preference.

Building such a benchmark platform poses its own practical challenges.
First, tasks must specify rich, diverse 3D worlds clearly while leaving agents free to build from an empty folder.
Second, because every agent invents its own codebase, automated evaluation cannot assume interfaces fixed in advance: rule checks have to be adapted to each game's code after it is built, and 3D visuals and control feel have to be judged from the game as it actually runs in a browser.
Third, judging whether a game is genuinely fun requires running live, blind head-to-head playtests and turning noisy human votes into reliable rankings.

\begin{figure*}[tp]
\centering
\input{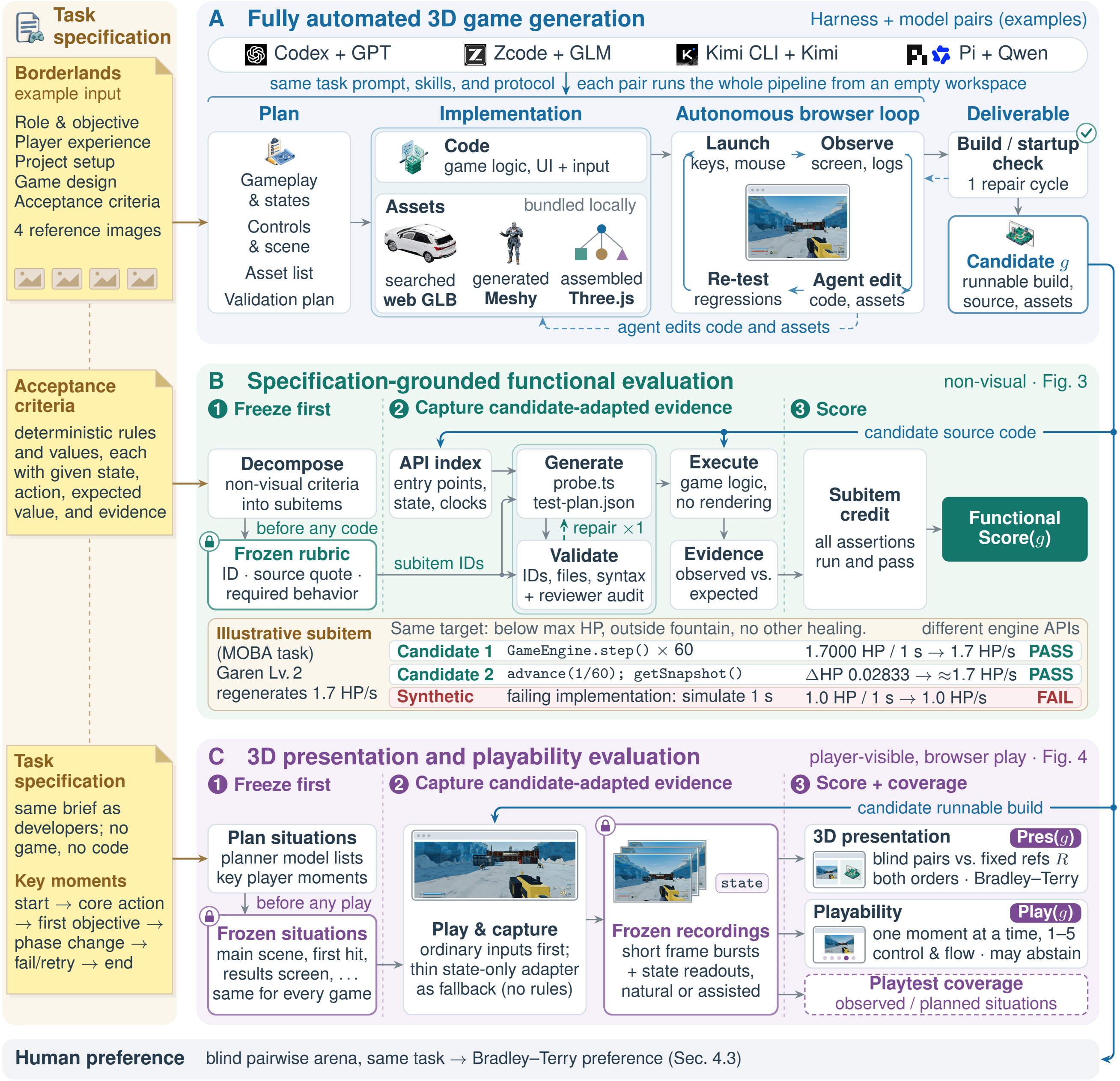}
\caption{\textbf{Overview of \bench{}.} \textbf{A:} a harness--model pair turns the task specification into a browser game. The resulting candidate is then evaluated by two independent tracks that share three steps: freeze the standard before any candidate is seen, capture candidate-adapted evidence, and score it. \textbf{B:} source-adapted tests check non-visual rules (Section~\ref{sec:code-acceptance}). \textbf{C:} recorded browser play is judged for 3D presentation and playability (Section~\ref{sec:player-experience}). The task note, harness--model pairs, and subitem rows are illustrative (Appendix~\ref{app:functional-details}).}
\label{fig:framework}
\end{figure*}

To address these gaps, we introduce \bench{} (Figure~\ref{fig:framework}), in which a fully automated pipeline carries an agent from planning through implementation, build, and autonomous playtest to a self-contained browser game. Our contributions are:
\begin{itemize}
    \item \textbf{A from-scratch, long-horizon 3D benchmark.} 108 tasks across 14 genres and two difficulty tiers, each a vertical slice of a popular game, given as a text specification with reference images. Agents get no starter code, bundled assets, or test hooks, and must find or generate their own 3D assets; a median build takes 2.6--7.9 hours and 692--1,067 tool calls (Section~\ref{sec:benchmark}).
    \item \textbf{Three-level evaluation, from code to player.} Functional tests keep the rubric fixed and adapt the test code to each game, so no hooks are needed during development; agent playtesting scores 3D presentation and playability from recorded browser play; blind pairwise human votes give a Bradley--Terry preference score (Section~\ref{sec:evaluation}).
    \item \textbf{Findings on what makes a preferred game.} The best of four configurations scores 64.7/100 functionally, and two configurations tied on functional score differ more than twofold in human preference. Development logs show that agents rarely lose track of the goal; the preferred ones test the running game, not just its code (Sections~\ref{sec:results}--\ref{sec:effort_and_assets}).
\end{itemize}

\section{Related Work}
\label{sec:related}
\paragraph{Game Generation Benchmarks.}
Game benchmarks for coding agents range from single-file 2D Pygame programs~\citep{zhang2026vgamegym} and localized edits inside existing Godot and Unreal projects~\citep{chi2026gamedevbench, la2026gameenginebench, che2026gamelogicbench} to end-to-end generation of whole games across engines and web runtimes~\citep{sun2026jamer, luo2026gamecraft, jiang2026opengame, chen2026gamexpert, gameasg2026}.
Compared with these (Table~\ref{tab:related}; Appendix~\ref{app:related-benchmarks}), \bench{} asks for full 3D games whose 3D models the agent must find or generate itself, starts from an empty, stack-agnostic workspace with no template or test interface, and adds blind human preference to rule compliance.

\paragraph{Evaluating Generated Games.}
Prior work checks generated games with white-box state and assertion tests~\citep{che2026gamelogicbench, gameasg2026} or with vision- and GUI-based playtesting~\citep{zhang2026vgamegym, huang2026playtestarena}; pairwise human preference with Bradley--Terry rankings~\citep{bradley1952rank, chiang2024chatbotarena, webdevarena2025} is standard for text and web generation but rare for 3D games.
We combine all three (Appendix~\ref{app:related-eval}).

\paragraph{Game Generation Agents.}
Systems range from multi-agent frameworks for lightweight 2D games~\citep{qian2024chatdev, hong2024metagpt} and coding agents with GUI playtesting~\citep{jiang2026opengame, playcoder2026} to Unity and Unreal pipelines for 3D~\citep{earle2025dreamgarden, yin2026autoue}.
\bench{} is tied to none of them and evaluates any coding agent (Appendix~\ref{app:related-agents}).

\begin{table*}[tp]
\caption{\textbf{Comparison with game-generation benchmarks:} task inputs, the length of an agent run, and evaluation.}
\label{tab:related}
\centering
\small
\setlength{\tabcolsep}{3.5pt}
\renewcommand{\arraystretch}{1.12}
\begin{tabularx}{\linewidth}{@{}lcc>{\raggedright\arraybackslash}p{1.95cm}>{\raggedright\arraybackslash}p{2.75cm}Y@{}}
\toprule
Benchmark & 3D/Web & Input & Starter code & Run length & Evaluation \\
\midrule
GameCraft~\citep{gamecraft2026} & \no/\no & T,A & NR & 128 tool calls; 2\,h cap & Replay + VLM rubric \\
GameXpert~\citep{gamexpert2026} & \yes/\yes & T,S & No (Gen);\newline Yes (Fix/Opt) & NR & Runtime tests, agent judge, human ratings \\
GameASG~\citep{gameasg2026} & \yes/\yes & T & No & 120-turn cap & Static + executable browser checks \\
GameGen-Verifier~\citep{jia2026gamegenverifier} & \nr/\yes & T & NR & NR & State injection, VLM, assertions \\
WebGameBench~\citep{webgamebench2026} & \nr/\yes & T & Template & 20--24 turns; 2\,h cap & Browser agent judge \\
SWE-Game~\citep{swegame2026} & \yes/\no & T,A,V,S & Varies by mode & 57--111 tool calls & Evaluator checks, reference comparison, VLM rubric \\
CEGame-Bench~\citep{cegame2026} & \yes/\no & T,A & Starter project & $\le$295 steps; 1\,h cap & Hidden checks, player journey, VLM judge \\
\midrule
\textbf{\bench{}} & \yes/\yes & \textbf{T,I} & \textbf{No} & \textbf{692--1,067 tool calls; 2.6--7.9\,h} & \textbf{Code tests}; agent playtesting; \textbf{human preference} \\
\bottomrule
\end{tabularx}
\vspace{3pt}
\begin{minipage}{\linewidth}
\footnotesize
T: text; S: source; A: assets; I: images; V: reference video. Inputs span task modes. 3D/Web: \yes{} yes, \no{} no, \nr{} unreported. Starter code means a supplied game implementation; evaluator-only references do not count. NR: not reported. Template: workspace scaffold, contents unspecified. Starter project: engine project with bundled assets and naming tables but no game logic.\par
\textbf{Run length} is the agent's effort on one task as each paper reports it; ``cap'' is an imposed limit. Ours: range of the per-configuration medians of tool calls and of build time (median: 881--1,980 tool calls; Table~\ref{tab:effort_main}).
\end{minipage}
\end{table*}

\section{Benchmark}
\label{sec:benchmark}
Tasks are drawn from widely played, top-charting commercial titles on Steam, online PC platforms, and console storefronts. A full commercial production, especially an AAA release, has sprawling progression systems and tens of hours of content, so we cut each large title into one to three self-contained \emph{vertical slices}. A slice isolates one signature gameplay mode at full mechanical depth: an action role-playing title inspired by \emph{Monster Hunter} becomes a single boss-arena fight, and an arcade racer inspired by \emph{Crazyracing KartRider} a single multi-lap circuit.

\subsection{Benchmark Construction}
\label{sec:benchmark-construction}

To turn a slice into an unambiguous engineering prompt that keeps the intended experience, we write its specification on Guardiola's \emph{Gameplay Loop} model~\citep{guardiola2016gameplay}, which describes a game as recurring cycles of player objectives, actions, state transitions, and feedback. Every specification runs 2,500--5,100 words and has the same five sections (Section~\ref{sec:task-specifications}).
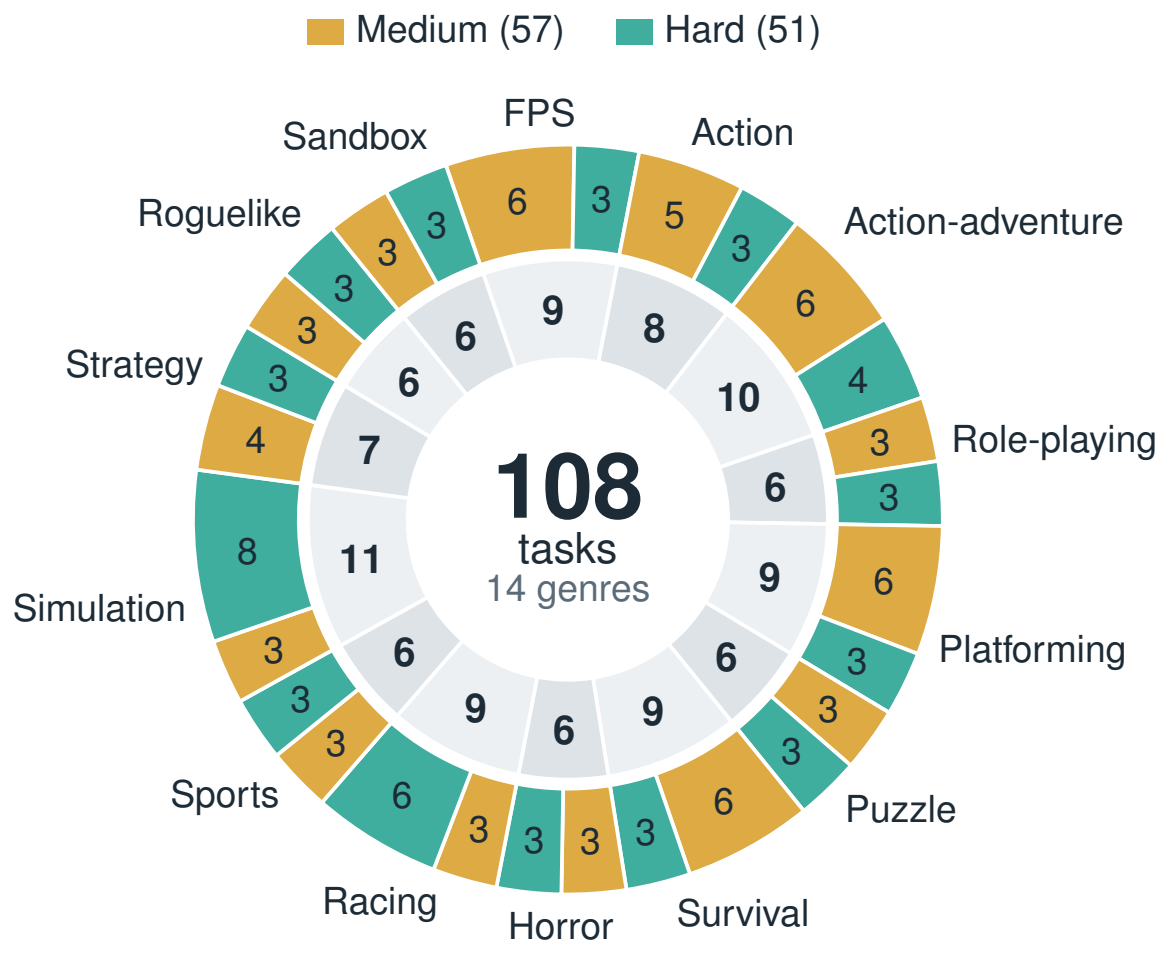
\begin{figure}[t]
\centering
\definecolor{tdInk}{HTML}{1C2B36}%
\definecolor{tdMuted}{HTML}{5B6B78}%
\definecolor{tdMed}{HTML}{DDAA44}%
\definecolor{tdHard}{HTML}{3FAE9F}%
\definecolor{tdRingA}{HTML}{EDF0F3}%
\definecolor{tdRingB}{HTML}{DEE3E8}%
\def\tdData{FPS/6/3, Action/5/3, Action-adventure/6/4, Role-playing/3/3, Platforming/6/3,
  Puzzle/3/3, Survival/6/3, Horror/3/3, Racing/3/6, Sports/3/3, Simulation/3/8, Strategy/4/3,
  Roguelike/3/3, Sandbox/3/3}%
\def\tdRa{1.06}\def\tdRb{1.72}\def\tdRc{1.78}\def\tdRd{2.48}%
\def\tdStart{109}%
\def\tdWedge#1#2#3#4#5{\path[fill=#5,draw=white,line width=.7pt,line join=round]
  (#3:#1) arc[start angle=#3,end angle=#4,radius=#1] -- (#4:#2)
  arc[start angle=#4,end angle=#3,radius=#2] -- cycle;}%
\def\tdFont#1#2{\fontfamily{phv}\selectfont\fontsize{#1}{#2}\selectfont}%
\begin{tikzpicture}[x=1cm,y=1cm,
  every node/.style={font=\tdFont{7}{8},text=tdInk,inner sep=0pt,outer sep=0pt}]
\xdef\tdTot{0}\xdef\tdM{0}\xdef\tdH{0}\xdef\tdN{0}%
\foreach \name/\m/\h in \tdData {%
  \pgfmathtruncatemacro\t{\tdTot+\m+\h}\xdef\tdTot{\t}%
  \pgfmathtruncatemacro\t{\tdM+\m}\xdef\tdM{\t}%
  \pgfmathtruncatemacro\t{\tdH+\h}\xdef\tdH{\t}%
  \pgfmathtruncatemacro\t{\tdN+1}\xdef\tdN{\t}}
\xdef\tdCum{0}%
\foreach \name/\m/\h [count=\i] in \tdData {%
  \pgfmathsetmacro\aS{\tdStart-\tdCum/\tdTot*360}%
  \pgfmathsetmacro\aM{\tdStart-(\tdCum+\m)/\tdTot*360}%
  \pgfmathsetmacro\aE{\tdStart-(\tdCum+\m+\h)/\tdTot*360}%
  \pgfmathsetmacro\aG{(\aS+\aE)/2}%
  \pgfmathsetmacro\aMm{(\aS+\aM)/2}%
  \pgfmathsetmacro\aHm{(\aM+\aE)/2}%
  \pgfinterruptpicture\setbox0=\hbox{\tdFont{7}{8}\name}\xdef\tdW{\the\wd0}\endpgfinterruptpicture
  \pgfmathsetmacro\tdRl{\tdRd+0.09+0.5*(\tdW)/28.45274*abs(cos(\aG))+0.11*abs(sin(\aG))}%
  \pgfmathtruncatemacro\tdSum{\m+\h}%
  \ifodd\i\def\tdFill{tdRingA}\else\def\tdFill{tdRingB}\fi
  \tdWedge{\tdRa}{\tdRb}{\aS}{\aE}{\tdFill}
  \tdWedge{\tdRc}{\tdRd}{\aS}{\aM}{tdMed}
  \tdWedge{\tdRc}{\tdRd}{\aM}{\aE}{tdHard}
  \node[font=\tdFont{7.5}{8.5}\bfseries] at (\aG:{(\tdRa+\tdRb)/2}) {\tdSum};
  \node[font=\tdFont{7}{8}] at (\aMm:{(\tdRc+\tdRd)/2}) {\m};
  \node[font=\tdFont{7}{8}] at (\aHm:{(\tdRc+\tdRd)/2}) {\h};
  \node at (\aG:\tdRl) {\name};
  \pgfmathtruncatemacro\t{\tdCum+\m+\h}\xdef\tdCum{\t}}
\node[font=\tdFont{17}{17}\bfseries] at (0,0.22) {\tdTot};
\node[font=\tdFont{8}{9}] at (0,-0.19) {tasks};
\node[font=\tdFont{7}{8},text=tdMuted] at (0,-0.48) {\tdN\ genres};
\fill[tdMed] (-1.72,\tdRd+0.66) rectangle ++(0.24,0.17);
\node[anchor=west] at (-1.40,\tdRd+0.745) {Medium (\tdM)};
\fill[tdHard] (0.32,\tdRd+0.66) rectangle ++(0.24,0.17);
\node[anchor=west] at (0.64,\tdRd+0.745) {Hard (\tdH)};
\end{tikzpicture}%
\caption{\textbf{Task distribution.} The inner ring gives the number of tasks in each of the 14 genres; the outer ring splits each genre into \emph{Medium} and \emph{Hard} tasks.}
\label{fig:benchmark-distribution}
\end{figure}

\paragraph{Coverage by Genre and Difficulty.}
The benchmark covers 14 canonical game genres at two implementation-difficulty tiers (Figure~\ref{fig:benchmark-distribution}). \emph{Medium} tasks are mostly 2D, 2.5D, and 3D games with bounded spatial dynamics and simple asset pipelines. \emph{Hard} tasks are all 3D vertical slices modeled on commercial and AAA productions; they demand complex spatial reasoning, skeletal and structural mesh integration, camera control, and real-time 3D physics. Every genre has at least three distinct tasks in each tier.

\subsection{Task Specifications}
\label{sec:task-specifications}

The five sections are (1)~\emph{Role and Objective}, (2)~\emph{Game Overview and Player Experience}, (3)~\emph{Project Setup and Deliverables}, (4)~\emph{Complete Game Design}, and (5)~\emph{Acceptance Criteria}; the fifth is written by a separate agent from the other four (Appendix~\ref{app:acceptance-criteria}).
\emph{Complete Game Design} ties player roles, camera perspectives, and input bindings to state variables, resource economies, entity interactions, the core gameplay loop~\citep{guardiola2016gameplay}, and terminal win, loss, and retry transitions.
Table~\ref{tab:task-specification} (Appendix~\ref{app:task-specification}) shows the five sections for two representative 3D tasks, a MOBA lane duel inspired by \emph{League of Legends} and an arcade racing circuit inspired by \emph{Crazyracing KartRider}. 


\subsection{Generation protocol}
\label{sec:protocol}

An agent starts from an empty workspace with a free choice of technology stack and only an \texttt{index.html} entry requirement. There are no test hooks: the specification prescribes no state-reporting interface and no place where tests must go. Every game is built in two stages, planning and execution (Figure~\ref{fig:framework}A), by a coding agent augmented with \emph{Agent Skills}~\citep{wang2024voyager, anthropic2025agentskills, xu2026agentskills, li2026skillsbench}: packages of workflows, verification scripts, and domain constraints loaded on demand. In the planning stage, a lead agent uses architectural-planning and game-orchestration skills~\citep{vincent2025superpowers, manzarpour2026threejs} to inspect the workspace contract and write an implementation blueprint covering core mechanics, state organization, controls, and 3D asset sourcing; no assets are supplied, so agents search the web for 3D models and generate missing ones with Meshy~\citep{meshy2026agent}. In the execution stage, an execution agent turns this blueprint into the playable deliverable with subagent-driven development and interactive browser-verification skills~\citep{vincent2025superpowers, openai2026developwebgame}, implementing game systems and playtesting until all acceptance criteria are met. Appendix~\ref{app:generation_protocol} gives the skill configurations, the 3D asset pipeline, and the exact prompts.
\vspace{-0.5em}
\paragraph{Evaluated configurations.}
We evaluate four configurations, each pairing a native coding-agent harness with one model at the highest reasoning setting of that harness: GPT-6.1 Sol and GPT-6 Astra in Codex CLI, GLM-5.3 in Zcode, and Kimi K3 in Kimi CLI (Table~\ref{tab:configs}, Appendix~\ref{app:generation_protocol}). Configuration IDs follow a harness--model pattern, such as \texttt{codex-gpt-6.1-sol}; because each model runs in exactly one harness, we refer to a configuration by its short name, such as \texttt{gpt-6.1-sol}.

\section{Evaluation}
\label{sec:evaluation}
We evaluate each game at three levels. Functional evaluation runs the game's code against the specification and checks whether it does what was asked (Section~\ref{sec:code-acceptance}; Figure~\ref{fig:framework}B). Agent playtesting scores 3D presentation and playability, that is, what a player sees and how the game feels to play (Section~\ref{sec:player-experience}; Figure~\ref{fig:framework}C). Finally, human players compare games in blind pairs, which tells us which games people actually prefer (Section~\ref{sec:preference}).

\subsection{Non-Visual Functional Evaluation}
\label{sec:code-acceptance}

Functional evaluation runs a game's own production logic against the requirements of its task, in three stages (Figure~\ref{fig:framework}B; worked example in Appendix~\ref{app:functional-details}). Its inputs are the candidate's registered source code and the acceptance criteria, which a separate agent derives from the rest of the task specification before any game is built (Figure ~\ref{app:acceptance-criteria}). The evaluation is non-visual: tests call the game's code or drive it with scripted inputs in a headless browser, and use no visible browser window, screenshot, or model judge. It thus verifies game-state logic, not what is visible or usable on screen.

\begin{figure*}[tp]
  \centering
  \includegraphics[width=\linewidth]{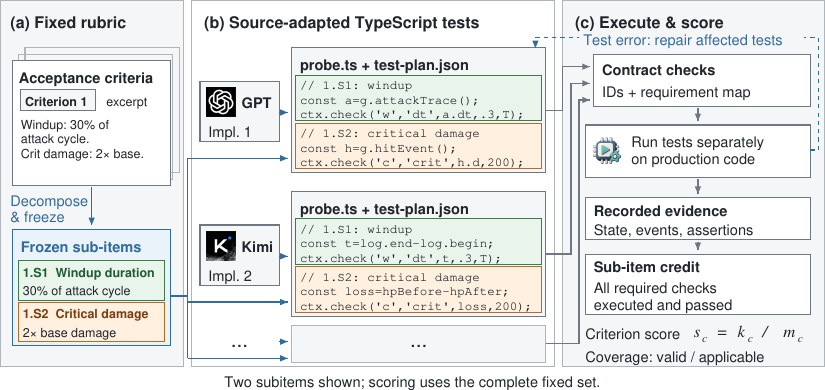}
  \caption{\textbf{Non-visual functional acceptance.} Gray probe cards show green windup (1.S1) and orange critical-damage (1.S2) excerpts for illustrative Criterion~1; full probes cover its frozen rubric. Schematic APIs run on separate fresh production fixtures (setup, actions, and remaining checks omitted). At 1\,Hz, 30\% windup is 0.3\,s under shared timing tolerance $T$: GPT reads recorded duration and damage, while Kimi uses accepted-start-to-release timestamps (seconds) and HP loss. Damage assumes base 100, ordinary $2\times$ critical hits, sufficient HP, and no mitigation, shields, healing, extra damage, or other modifiers (GPT/Kimi denote illustrative origins). Credit requires all subitem assertions to execute and pass, and repairs modify tests only.}
  \label{fig:acceptance-harness}
\end{figure*}

\paragraph{Stage 1: Freeze the rubric.}
Before any candidate code is seen, a spec reader splits each acceptance criterion $c$ into $m_c$ subitems and freezes them, keeping the subitem IDs, the verbatim requirement text, and the behaviors that must hold. Every candidate for a task is judged against this same rubric.

\paragraph{Stage 2: Adapt and run the tests.}
Tests are written after the build is finished and run on a copy of the delivered game. Test writers may attach debug interfaces to the copy that expose existing state but never change game functions, rules, or state; preconditions are reached through the game's own operations, unlike in playtesting (Section~\ref{sec:player-experience}). Candidates implement the same requirement behind different interfaces, so tests are written per candidate. Test writers inspect the candidate's interfaces, state, clocks, and entry points, and several independent writers produce unit tests for each subitem. A $30\%$ attack windup, for example, is checked through a duration field in one game and through start-to-release timestamps in another. A separate reviewer audits the tests and may revise them once; it may change test and driver code only, never the game code or the rubric. An isolated worker then runs every assertion against the real production functions on fresh fixtures. Expected values come from the specification alone, never from the candidate's own output or formula, so real functional failures stay failures; infrastructure errors are logged separately.

\paragraph{Stage 3: Score.}
A subitem passes only if all of its assertions execute and pass, and each passed subitem earns an equal $1/m_c$ share of its criterion, however many assertions it has. Let $V_g$ be the criteria of game $g$ that were validly evaluated and $k_{g,c}$ the number of passed subitems of criterion $c$. Criteria carry equal weight:
\begin{equation}
  s_{g,c} = \frac{k_{g,c}}{m_c}, \qquad
  \operatorname{FunctionalScore}(g) = \frac{100}{|V_g|}\sum_{c\in V_g}s_{g,c}.
\label{eq:functional-score}
\end{equation}
Criteria that could not be evaluated, for instance because of a test or infrastructure error, are left out of $V_g$ and are not counted as failures (Appendix~\ref{app:functional-details}). If $V_g$ is empty the score is undefined, not zero. A configuration's score is the mean over its games with a defined score, so real zeros stay in the mean.

\subsection{3D Presentation and Playability}
\label{sec:agent-evaluation}
\label{sec:player-experience}
A game can pass every functional test and still be hard to look at or hard to play, so we also score what the player experiences, on two axes. \emph{3D presentation} asks whether the rendered world lets the player see what is happening and what matters. It is a holistic quality that is easier to compare than to rate, so we score it pairwise. \emph{Playability} asks whether actions get clear responses and whether the player knows where the game stands. It is local, so we rate it one moment at a time (Appendix Figure~\ref{fig:evaluator}).

\paragraph{Situations.}
A planner model reads only the task specification, the brief the developers received, and lists the key \emph{situations} a player meets in ordinary play, such as entering the main scene or reaching the results screen, each with what the player should be able to notice. The plan is fixed before any game is played, so all games for a task are judged on the same moments.

\paragraph{Recording.}
An executor agent plays each game in the browser and records a short burst of frames and state readouts at each situation. It uses ordinary inputs where it can and otherwise a thin evaluation-time adapter that reads or sets the game's own state but never implements game rules. Each moment is labeled \emph{natural} or \emph{assisted} (Appendix~\ref{sec:player-experience-arrival}); an assisted moment shows the experience in that state, not that play reaches it. All judging uses only these frozen recordings.

\paragraph{3D presentation.}
A small fixed set $R$ of reference games, spanning the quality range, sits on a common Bradley--Terry scale with strengths $r_a$. A judge model compares the recordings of game $g$ with those of each reference side by side, with identities hidden, and grades each outcome. We fit the strength $\theta_g$ with the references held fixed and report the expected win rate against them (details in Appendix~\ref{sec:presentation-scoring}):
\begin{equation}
\operatorname{Pres}(g)=\frac{100}{|R|}\sum_{a\in R}\sigma(\theta_g-r_a).
\label{eq:presentation}
\end{equation}
Because the references never change, scores are comparable across tasks and do not depend on which other games are in the pool.

\paragraph{Playability.}
Playability covers control, whether an action gets a clear and timely response, and flow, whether the player knows the current state and the next step at turning points such as onboarding, failure, or the ending. The judge sees one situation at a time, never another game, and picks a level on a rubric written in terms of the player's experience (Appendix~\ref{sec:playability-rubric}), or abstains when the frames are not enough. With $\ell_m\in\{1,\dots,5\}$ the level of moment $m$ and $M_{g,d}$ the rated moments of game $g$ for aspect $d$,
\begin{equation}
\operatorname{Play}(g)=\frac{100}{|D_g|}\sum_{d\in D_g}\frac{1}{|M_{g,d}|}\sum_{m\in M_{g,d}}\frac{\ell_m-1}{4},
\label{eq:playability}
\end{equation}
where $D_g\subseteq\{\text{control},\text{flow}\}$ are the aspects with at least one rated moment. Situations that cannot be observed in a game are left out of both scores, and we report the observed fraction as Coverage.

\subsection{Human Preference}
\label{sec:preference}
In a public arena, a participant plays two games that different configurations built for the same task and votes for the better one. Presentation order is randomized, and configuration identities stay hidden until the vote is stored. A vote is a preference for either game, a tie, or both bad; ties and both-bad votes count as half a win for each side, and skipped comparisons are discarded.

We fit a Bradley--Terry model with one strength $\beta_a$ per configuration $a$, shared across tasks, and report the \emph{BT Score}, the expected win rate of a configuration against an average opponent:
\begin{equation}
\begin{aligned}
 P(a\succ b)&=\sigma(\beta_a-\beta_b),\\
 \operatorname{Preference}(a)&=\frac{100}{K-1}\sum_{b\ne a}\sigma(\beta_a-\beta_b),
\end{aligned}
 \label{eq:human-preference}
\end{equation}
where $\sigma$ is the logistic function and $K$ is the number of configurations. The BT Score adjusts for the strength of the opponents a configuration actually met. The strengths are fitted by regularized maximum likelihood, with each vote weighted by the inverse probability that its matchup was sampled. We also report \emph{Raw Win}, the plain share of a configuration's comparisons that it won, which makes no such adjustment. Appendix~\ref{app:preference-details} gives the likelihood, the weights, and the Raw Win formula.

\section{Results}
\label{sec:results}

Table~\ref{tab:main_results} reports the three evaluation levels of Section~\ref{sec:evaluation} for each configuration. We draw three conclusions:

\begin{itemize}
    \item \textbf{From-scratch 3D game development is far from solved.} Even the best configuration earns under two-thirds of the functional score, and no configuration reaches $70$ on either agent-playtesting score.
    
    \item \textbf{Correct code does not make a preferred game.} \texttt{kimi-k3} and \texttt{gpt-6-astra} are tied on the functional score, yet the BT score of \texttt{gpt-6-astra} is more than twice that of \texttt{kimi-k3}. Functional tests alone would rank the two as equals.
    
    \item \textbf{The closer a measure is to the player, the wider the gap.} The best and worst configurations are less than $15$ points apart on the functional score and on playability, $35$ points apart on 3D presentation, and more than $40$ points apart in human preference. Agent playtesting sees the visual gap that functional tests miss, but it ranks \texttt{kimi-k3} well above \texttt{glm-5.3}, and players do not. The three levels are therefore complementary, not interchangeable.
\end{itemize}

Where do these gaps come from? To find out, we go back to the development logs and analyze how the agents work: where their effort goes, how they debug, how they use 3D assets, and whether they stay on task over a long run. Section~\ref{sec:effort_and_assets} reports this analysis as a series of takeaways.

\begin{table*}[t]
\centering
\caption{\textbf{Main results.} Each column group is one evaluation level of Section~\ref{sec:evaluation}. \textbf{Functional} reports the functional score from non-visual tests (Section~\ref{sec:code-acceptance}). \textbf{Agent playtesting} reports 3D presentation and playability, and the share of planned situations that were observed (Coverage; Section~\ref{sec:player-experience}). \textbf{Human preference} reports the Bradley--Terry expected win rate ($\times 100$) against an average opponent (BT Score) and the empirical pairwise win rate (Raw Win; Section~\ref{sec:preference}). All values are on a $0$--$100$ scale. Best score per column in bold.}
\label{tab:main_results}
\small
\setlength{\tabcolsep}{6pt}
\begin{tabular}{lcccccc}
\toprule
 & \textbf{Functional} & \multicolumn{3}{c}{\textbf{Agent playtesting}} & \multicolumn{2}{c}{\textbf{Human preference}} \\
\cmidrule(lr){2-2}\cmidrule(lr){3-5}\cmidrule(lr){6-7}
\textbf{Configuration} & Score $\uparrow$ & 3D Presentation $\uparrow$ & Playability $\uparrow$ & Coverage & BT Score $\uparrow$ & Raw Win $\uparrow$ \\
\midrule
\texttt{codex-gpt-6.1-sol} & \textbf{64.7} & 57.8 & 68.4 & 91.3 & 69.11 & 62.8 \\
\texttt{codex-gpt-6-astra} & 59.4 & \textbf{58.5} & \textbf{69.5} & 95.8 & \textbf{69.24} & \textbf{66.2} \\
\texttt{kimi-k3} & 59.8 & 42.3 & 64.9 & 95.9 & 25.86 & 27.4 \\
\texttt{zcode-glm-5.3} & 51.2 & 23.5 & 54.9 & 93.2 & 35.78 & 30.6 \\
\bottomrule
\end{tabular}
\end{table*}

\section{Analysis}
\label{sec:effort_and_assets}

We present the analysis as takeaways: each states one finding in a sentence and supports it with a short paragraph, with the full numbers in the tables. The takeaways answer four questions: where the engineering effort goes and what it buys (Section~\ref{sec:effort_roi}), how agents use 3D assets (Section~\ref{sec:asset_strategy}), what agent playtesting does and does not capture (Section~\ref{sec:presentation_playability}), and whether agents stay on task over a long run (Section~\ref{sec:long_horizon}).

\begin{table*}[t]
\centering
\caption{\textbf{Long-horizon engineering effort, codebase size, and human Arena preference.} Values report \textbf{Median $[\text{Q1}, \text{Q3}]$}. Using measured $8\text{K}$-input end-to-end total output speeds ($26.93$--$33.33\text{ tok/s}$), \textbf{Norm.\ Time} normalizes active execution duration to a uniform $30\text{ tok/s}$ output rate. \textbf{Clean LOC} counts handwritten source lines of code after removing blank lines, comments, build bundles, and third-party libraries. \textbf{BT Score} reports the weighted Bradley--Terry expected win rate ($\times 100$) against an average opponent, alongside empirical \textbf{Raw Win} rate.}
\label{tab:effort_main}
\resizebox{\textwidth}{!}{%
\begin{tabular}{lccccccccc}
\toprule
\textbf{Configuration} & \textbf{Norm.\ Time} (h) & $\textbf{Tok}_{\textbf{total}}$ ($10^6$) & $\textbf{Tok}_{\textbf{out}}$ ($10^6$) & \textbf{Tool Calls} & \textbf{Turns / Sub-tasks} & \textbf{Source Files} & \textbf{Clean LOC} & \textbf{BT Score} & \textbf{Raw Win} \\
\midrule
\textbf{\texttt{codex-gpt-6.1-sol}} & \textbf{3.82} $[1.98, 8.29]$ & \textbf{162.0} $[84.2, 317.5]$ & \textbf{0.624} $[0.391, 1.222]$ & \textbf{1,067} $[595, 2,016]$ & \textbf{1,186} / \textbf{18.5} & \textbf{35.0} $[25.5, 53.0]$ & \textbf{2,328} $[1,523, 4,666]$ & 69.11 & 62.8\% \\
\textbf{\texttt{codex-gpt-6-astra}} & \textbf{2.72} $[1.86, 5.48]$ & \textbf{121.5} $[77.7, 212.9]$ & \textbf{0.532} $[0.346, 0.853]$ & \textbf{921} $[602, 1,647]$ & \textbf{1,044} / \textbf{17.0} & \textbf{39.5} $[32.2, 54.5]$ & \textbf{2,885} $[2,151, 4,046]$ & \textbf{69.24} & \textbf{66.2\%} \\
\textbf{\texttt{zcode-glm-5.3}} & \textbf{2.65} $[1.75, 5.01]$ & \textbf{70.1} $[40.1, 104.0]$ & \textbf{0.456} $[0.259, 0.733]$ & \textbf{692} $[426, 1,151]$ & \textbf{645} / \textbf{25.0} & \textbf{82.0} $[57.0, 122.0]$ & \textbf{13,824} $[8,603, 20,406]$ & 35.78 & 30.6\% \\
\textbf{\texttt{kimi-k3}} & \textbf{7.11} $[4.02, 11.36]$ & \textbf{202.1} $[91.7, 305.4]$ & \textbf{0.694} $[0.369, 1.572]$ & \textbf{1,045} $[529, 2,115]$ & \textbf{756} / \textbf{7.0} & \textbf{51.5} $[38.5, 81.2]$ & \textbf{7,572} $[4,644, 13,564]$ & 25.86 & 27.4\% \\
\bottomrule
\end{tabular}%
}
\end{table*}

\subsection{Engineering Effort, Debugging, and Preference}
\label{sec:effort_roi}

\takeaway{Takeaway 1}{Writing more code does not make a better game; top models keep codebases small and modular.}
The two configurations that players prefer leave the smallest codebases, with medians under $3{,}000$ clean lines of code (Table~\ref{tab:effort_main}). \texttt{kimi-k3} and \texttt{glm-5.3} write roughly three to six times as much, in files at least twice as long. Large files make it harder for an agent to keep track of state shared across modules, so later edits tend to break earlier features.


\begin{table}[!htbp]
\centering
\caption{\textbf{Token breakdown and human preference.} Per-game means. \textbf{Uncached In} is new text entering the context, \textbf{Cache Read} is history re-read from the cache, and \textbf{Cache Replay} is their ratio.}
\label{tab:token_decomp}
\resizebox{0.8\linewidth}{!}{%
\begin{tabular}{lcccc}
\toprule
\textbf{Config.} & \textbf{BT Score} & \textbf{Uncached In} & \textbf{Cache Read} & \textbf{Cache Replay} \\
\midrule
\texttt{gpt-6.1-sol} & 69.11 & 6.82M & 307.41M & $45.1\times$ \\
\texttt{gpt-6-astra}   & \textbf{69.24} & 5.15M & 200.46M & $38.9\times$ \\
\texttt{kimi-k3}     & 25.86 & 4.91M & 241.04M & $49.1\times$ \\
\texttt{glm-5.3}     & 35.78 & 1.86M & 82.27M  & $44.3\times$ \\
\bottomrule
\end{tabular}%
}
\end{table}

\takeaway{Takeaway 2}{Almost all input tokens are cached context re-read at every tool call; a runaway total signals a stalled build, not progress.}
New text entering the context (file reads, compiler errors, browser logs) stays at a few million tokens per build. Totals reach tens to hundreds of millions because every tool call re-reads the project history from the prefix cache, $39$--$49$ times over (Table~\ref{tab:token_decomp}), so totals track the number of tool calls and cache discounts keep long builds affordable. More is not better: within every configuration, builds that become playable use $20\%$--$30\%$ fewer tool calls than builds that stall. Splitting work into sub-tasks keeps each history short; \texttt{kimi-k3} does this least, and even at equal output speed its median build takes roughly two to two and a half times as long as those of the Codex configurations (Table~\ref{tab:effort_main}).

\takeaway{Takeaway 3}{Extra tokens buy higher Arena preference only when spent on live testing.}
More tokens alone do not make a preferred game: \texttt{kimi-k3} spends about three times as many as \texttt{glm-5.3} and is preferred less (Tables~\ref{tab:effort_main} and~\ref{tab:token_decomp}). The Codex configurations, which players prefer, also spend more than \texttt{glm-5.3}, but they debug mostly by live testing, with full builds and browser checks, where \texttt{kimi-k3} mostly runs static checks that never render a frame (Table~\ref{tab:debug_mix}). Even there the return flattens: \texttt{gpt-6.1-sol} outspends \texttt{gpt-6-astra} by a third to a half, and players rate the two alike.

\begin{table}[!htbp]
\centering
\caption{\textbf{Debugging and outcomes.} Debug Calls: median (mean) per game, split into static and unit checks (\texttt{tsc}+\texttt{vitest}), project \texttt{build}+\texttt{test}, and live browser smoke tests. }
\label{tab:debug_mix}
\resizebox{\linewidth}{!}{%
\begin{tabular}{lcccccc}
\toprule
\textbf{Config.} & \textbf{Debug Calls} & \textbf{\texttt{tsc}+\texttt{vitest}} & \textbf{\texttt{build}+\texttt{test}} & \textbf{Smoke} & \textbf{Func.\ Score} & \textbf{BT / Raw Win} \\
\midrule
\texttt{gpt-6.1-sol} & 103.0 {\scriptsize(184.6)} & 8.4\%  & \textbf{83.3\%} & 8.1\%  & \textbf{64.7\%} & 69.1 / 62.8\% \\
\texttt{gpt-6-astra}   & 55.0 {\scriptsize(120.9)}  & 11.2\% & 76.8\%          & 11.4\% & 59.4\% & \textbf{69.2} / \textbf{66.2\%} \\
\texttt{kimi-k3}     & \textbf{159.5} {\scriptsize(167.9)} & \textbf{63.5\%} & 17.7\% & 16.9\% & 59.8\% & 25.9 / 27.4\% \\
\texttt{glm-5.3}     &  103.5 {\scriptsize(120.1)} & 63.4\% & 15.2\% & \textbf{21.1\%} & 51.2\% & 35.8 / 30.6\% \\
\bottomrule
\end{tabular}%
}
\end{table}

\takeaway{Takeaway 4}{What an agent debugs matters more than how often it debugs: static unit tests raise functional scores, whereas runtime testing wins human preference.}
Two comparisons show why functional scores and player preference diverge (Table~\ref{tab:debug_mix}). In the same harness, more debugging raises the functional score but not preference: \texttt{gpt-6.1-sol} runs nearly twice as many debug commands per game as \texttt{gpt-6-astra} and scores higher on the functional tests, yet players rate the two alike. Across harnesses, the kind of debugging decides. \texttt{kimi-k3} has the highest median debug count and matches \texttt{gpt-6-astra} on the functional score, yet its BT score is well under half as high. It spends $63.5\%$ of its debugging on type checks and isolated unit tests, which pass without rendering a single frame. The two Codex configurations spend about $90\%$ on full builds, integration tests, and live browser checks, which is where broken hitboxes, camera framing, and controls show up.

\subsection{3D Asset Strategies and Player Enjoyment}
\label{sec:asset_strategy}

\begin{table}[!htbp]
\centering
\caption{\textbf{3D asset strategies in the playable builds we inspected.} Builds rely on built-in primitives only (Prim), retrieved external models only (Ret), AI-generated models only (Gen), hybrid external models (Hyb), or dynamic loaders (Dyn): $\bullet$ at least a quarter of builds, $\circ$ fewer, --- none.}
\label{tab:asset_strategy}
\resizebox{\linewidth}{!}{%
\begin{tabular}{lcccccccc}
\toprule
 & \multicolumn{5}{c}{\textbf{Strategy}} & \multicolumn{3}{c}{\textbf{Mean count per game}} \\
\cmidrule(lr){2-6}\cmidrule(lr){7-9}
\textbf{Config.} & \textbf{Prim} & \textbf{Ret} & \textbf{Gen} & \textbf{Hyb} & \textbf{Dyn} & \textbf{Generated} & \textbf{Retrieved} & \textbf{Primitives} \\
\midrule
\texttt{gpt-6-astra} & $\bullet$ & $\circ$ & $\circ$ & $\circ$ & $\bullet$ & $<1$ & $<1$ & 17 \\
\texttt{gpt-6.1-sol} & $\bullet$ & $\bullet$ & --- & --- & $\bullet$ & 0 & $<1$ & 8 \\
\texttt{kimi-k3}     & --- & $\bullet$ & --- & --- & --- & 0 & 13 & 60 \\
\texttt{glm-5.3}     & $\bullet$ & --- & --- & --- & --- & 0 & 0 & 11 \\
\bottomrule
\end{tabular}%
}
\end{table}

\takeaway{Takeaway 5}{Adding more 3D objects does not make a better game.}
\texttt{kimi-k3} puts by far the most 3D content into its games, on average $13$ downloaded models and $60$ primitives per game, yet falls in the lower tier of human preference (Tables~\ref{tab:main_results} and~\ref{tab:asset_strategy}). In agent playtesting, its games fall short on 3D presentation rather than on playability (Takeaway~7). A likely reason is that downloaded meshes arrive with their own scale, orientation, and framing, which the agent must reconcile with the rest of the scene.

\takeaway{Takeaway 6}{Wiring AI-generated 3D models into live gameplay remains an open bottleneck for coding agents.}
Even with 3D generation tools available, only \texttt{gpt-6-astra} integrates AI-generated 3D models into playable final builds, and only in a small share of them; no other configuration has a generated mesh in working runtime code (Table~\ref{tab:asset_strategy}). Calling an API to create a static mesh is straightforward, but autonomously checking its proportions, pivot point, and materials and rigging it into interactive game logic remains largely unsolved.

\subsection{3D Presentation and Playability}
\label{sec:presentation_playability}


\takeaway{Takeaway 7}{3D presentation separates agents far more than moment-level playability, and neither captures a whole play session.}
Playability spans only about $15$ points across configurations, also when only naturally reached moments are counted, because at the planned key moments most games already give readable feedback and a clear next step (Appendix~\ref{sec:player-experience-arrival}). 3D presentation spans $35$ points: \texttt{kimi-k3} nearly matches the top configurations on playability but trails by $16$ points on presentation, consistent with its reliance on downloaded assets (Takeaway~5), and most \texttt{glm-5.3} games sit at the bottom of the scale (Appendix Table~\ref{tab:presentation_playability}). Neither score separates the two Codex configurations (Appendix Table~\ref{tab:px-pairs}), and neither do players. Both scores put \texttt{kimi-k3} above \texttt{glm-5.3}, which players do not: our recordings are short silent clips of key moments, whereas Arena players play a whole session.

\takeaway{Takeaway 8}{Agents build clear menus and stage transitions far more reliably than responsive 3D action controls.}
Flow, whether prompts, menus, and stage transitions make the state and next goal clear, beats Control, whether an input gets immediate, readable feedback, in $86\%$ of all games and by a similar margin in every configuration (Appendix Table~\ref{tab:presentation_playability}). The two scales are not calibrated against each other, so the consistency across configurations is the more robust signal. About half of the \texttt{glm-5.3} games score below $3$ on Control, where a player can no longer tell whether an input registered. A likely reason is that flow is built from 2D overlays and state machines familiar from web development, whereas clear control needs input handling, camera framing, sightlines, and hit feedback to work together in real time.

\subsection{Long-Horizon Goal Retention and Plan Adaptation}
\label{sec:long_horizon}

A build runs for hours and around a thousand model calls, and the agent's context is compacted several times on the way, so we check whether a long run stays on its goal. We go back through the development logs, cut each run into four quarters (Q1--Q4), sample operations from each, and have an LLM reviewer answer two questions about every sampled operation. \emph{Goal retention} (GR) asks whether the operation still serves the task, including any change approved along the way~\citep{arike2025goaldrift,jia2025agentgpa}. \emph{Plan adherence} (PA) asks whether it follows the plan the agent made before it started coding~\citep{jia2025agentgpa,liu2026plantoaction}. Both measure whether the agent is still working on the right thing, not whether the work succeeds. Table~\ref{tab:long_horizon} also reports the largest tasks separately, where runs are longest (details in Appendix~\ref{sec:retrospective-appendix}).
\begin{table}[!htbp]
\centering
\caption{\textbf{Goal retention (GR) and plan adherence (PA) over a run.} \emph{Top:} large tasks, by quarter of the run (Q1 is the first, Q4 the last). \emph{Bottom:} all builds, with full-run GR as judged and with process-rule violations left aside (GR$^{\dagger}$), median model calls per build (including sub-agents) by task size, and BT score. }
\label{tab:long_horizon}
\resizebox{\linewidth}{!}{%
\begin{tabular}{@{}c@{}}
\begin{tabular}{lcccccccc}
\toprule
 & \multicolumn{4}{c}{\textbf{Goal retention (\%)}} & \multicolumn{4}{c}{\textbf{Plan adherence (\%)}} \\
\cmidrule(lr){2-5}\cmidrule(lr){6-9}
\textbf{Config.} & \textbf{Q1} & \textbf{Q2} & \textbf{Q3} & \textbf{Q4} & \textbf{Q1} & \textbf{Q2} & \textbf{Q3} & \textbf{Q4} \\
\midrule
\texttt{gpt-6.1-sol} & 100 & 100 & 100 & 100 & 100 & 100 & 96.2 & 91.4 \\
\texttt{gpt-6-astra} & 100 & 100 & 100 & 100 & 100 & 100 & 100 & 100 \\
\texttt{kimi-k3}     & 84.5 & 94.0 & 98.8 & 96.2 & 100 & 100 & 100 & 98.6 \\
\bottomrule
\end{tabular}\\[7pt]
\begin{tabular}{lcccccc}
\toprule
 & \multicolumn{2}{c}{\textbf{Full run (\%)}} & \multicolumn{3}{c}{\textbf{Calls by task size}} & \\
\cmidrule(lr){2-3}\cmidrule(lr){4-6}
\textbf{Config.} & \textbf{GR} & \textbf{GR$^{\dagger}$} & \textbf{Small} & \textbf{Medium} & \textbf{Large} & \textbf{BT} \\
\midrule
\texttt{gpt-6.1-sol} & 99.7 & 99.9 & 849 & 1,186 & 1,645 & 69.1 \\
\texttt{gpt-6-astra} & 100 & 100 & 971 & 1,132 & 1,404 & \textbf{69.2} \\
\texttt{kimi-k3}     & 96.1 & 99.3 & 601 & 730 & 761 & 25.9 \\
\bottomrule
\end{tabular}
\end{tabular}%
}
\end{table}

\takeaway{Takeaway 9}{Agents do not forget the game: late in a long run, they are still working on what was asked.}
No configuration drifts. Goal retention is between $96\%$ and $100\%$ over full runs and stays at $94\%$ or above on the largest tasks; where it dips on those tasks, as for \texttt{kimi-k3}, the dip comes in the first quarter of a run, not the last (Table~\ref{tab:long_horizon}; Appendix Table~\ref{tab:retrospective-sensitivity}). Forgetting the goal is therefore not what separates good builds from bad ones: \texttt{kimi-k3} stays on task almost as well as the Codex configurations, yet players rarely prefer its games. What differs is how the work is checked in the running game .

\takeaway{Takeaway 10}{When agents do step off the path, they leave the plan or break a process rule; they do not abandon the game.}
Two things do change. First, agents follow their initial plan less closely near the end: on large tasks, the plan adherence of \texttt{gpt-6.1-sol} falls from $100\%$ to $91.4\%$ in the last quarter while its goal retention stays at $100\%$. Nearly all of this off-plan work still serves the goal, typically scope approved after the plan was written, so it is adaptation, not drift, and a long run is better judged against its current goal than its first plan. Second, the few real violations concern how the agent was told to work, such as starting to code when asked only to plan, not what to build. Setting them aside raises the goal retention of \texttt{kimi-k3} from $96.1\%$ to $99.3\%$.

\section{Conclusion}
\label{sec:conclusion}
We introduced \bench{}, a benchmark of 108 tasks for long-horizon, from-scratch 3D game development, evaluated at three levels: functional tests, agent playtesting, and blind human preference. The task is far from solved: the best configuration earns under two-thirds of the functional score, and correct code does not make a game that players prefer. Our analysis traces this gap to how agents check their work, not to how well they remember it. Agents do not forget what they are building, even late in runs that last hours. But those that debug mainly with type checks and isolated unit tests pass functional tests without producing a game people want to play, while the preferred agents spend their debugging on full builds and live browser checks. The main limitation of \bench{} is the time its evaluation takes: every game needs its own adapted tests, a recorded play session, and human votes, which limits how many agents and repeated runs can be evaluated; making it faster is the natural next step.
\clearpage
\section*{LLM Usage Statement}
We used an LLM-based assistant (Claude, Anthropic) to polish  the manuscript.
All technical content, experiments, results, and claims are the authors' own, and all edited text was reviewed by the authors.
{\small
\bibliographystyle{ieeenat_fullname}
\bibliography{references}
}
\clearpage
\appendix
\onecolumn
\section{Extended Related Work}
\label{app:related-work}

This appendix gives the full discussion behind the three shortened paragraphs in Section~\ref{sec:related}.

\subsection{Game Generation Benchmarks}
\label{app:related-benchmarks}
Recent benchmarks have moved beyond generic software engineering testbeds~\citep{jimenez2024swebench, liu2025webgenbench} to evaluate coding agents on game creation.
Early efforts such as V-GameGym~\citep{zhang2026vgamegym} focus on single-file 2D Pygame synthesis, whereas engine-grounded benchmarks---including GameDevBench~\citep{chi2026gamedevbench}, GameEngineBench~\citep{la2026gameenginebench}, and GameLogicBench~\citep{che2026gamelogicbench}---assess localized feature implementation or tutorial-derived edits inside pre-structured Godot and Unreal repositories.
At the full-project scale, JAMER~\citep{sun2026jamer}, GameCraft-Bench~\citep{luo2026gamecraft}, OpenGame-Bench~\citep{jiang2026opengame}, GameXpert-Bench~\citep{chen2026gamexpert}, and GameASG-Bench~\citep{gameasg2026} evaluate end-to-end game generation across game engines and web runtimes.
\bench{} differs from these benchmarks in three respects: (i) whereas existing suites predominantly target 2D games or provide pre-bundled project assets, we target full-scale 3D games where agents must autonomously retrieve or synthesize dozens of heterogeneous 3D models from scratch; (ii) rather than constraining agents to fixed engine templates, localized edits, or pre-declared test interfaces~\citep{gameasg2026}, we enforce a zero-scaffolding, stack-agnostic workspace over multi-hour horizons; and (iii) beyond rule compliance, we establish blind human A/B playtesting to measure player preference. Table~\ref{tab:related} compares design choices and is not a performance ranking.

\subsection{Evaluating Generated Games}
\label{app:related-eval}
Assessing generated games requires verifying dynamic, stateful interactions that static code inspection and compilation checks cannot capture. 
Prior work has explored three main verification paradigms. 
First, \emph{white-box state and assertion testing} checks runtime invariants directly, ranging from tick-level engine assertions~\citep{che2026gamelogicbench}, runtime state injection~\citep{jia2026gamegenverifier}, and replayable demonstration traces~\citep{luo2026gamecraft} to pre-declared snapshot contracts~\citep{gameasg2026}. 
Second, \emph{vision- and GUI-based playtesting} builds on automated game QA~\citep{zheng2019wuji} by deploying multimodal judges and browser/GUI agents to interact with rendered frames and adjudicate behavioral rubrics, as in V-GameGym~\citep{zhang2026vgamegym}, OpenGame-Bench~\citep{jiang2026opengame}, PlayEval~\citep{playcoder2026}, and PlaytestArena~\citep{huang2026playtestarena}. 
Third, \emph{pairwise human preference modeling} via Bradley--Terry rankings~\citep{bradley1952rank} has become standard for open-ended text and web generation~\citep{chiang2024chatbotarena, webdevarena2025}, yet remains largely absent from 3D game benchmarks. 
Our evaluation pipeline bridges these paradigms by combining non-intrusive evaluation-time state adapters, automated browser playtesting agents, and blind human A/B preference rankings.

\subsection{Game Generation Agents}
\label{app:related-agents}
Early multi-agent software frameworks such as ChatDev~\citep{qian2024chatdev}, MetaGPT~\citep{hong2024metagpt}, and GameGPT~\citep{chen2023gamegpt} simulate studio roles to translate natural-language prompts into lightweight 2D games, but frequently suffer from cross-file inconsistencies and asset hallucinations at scale. 
Recent domain-specialized coding agents address long-horizon coherence through structured workflows and feedback loops: OpenGame~\citep{jiang2026opengame} pairs a specialized code model with reusable template and debugging skills for 2D web games, while PlayCoder~\citep{playcoder2026} and Play2Code~\citep{huang2026playtestarena} couple code-writing agents with GUI playtesters for iterative refinement. 
In 3D environments, world-model approaches synthesize interactive video frames directly~\citep{bruce2024genie, che2025gamegenx} without explicit software engineering, whereas engine-centric multi-agent pipelines---such as DreamGarden~\citep{earle2025dreamgarden}, UniGen~\citep{yang2025unigen}, and AutoUE~\citep{yin2026autoue}---orchestrate asset retrieval, procedural scene layout, and C++/C\# scripting inside Unity or Unreal Engine. 
Rather than tying evaluation to a single bespoke pipeline, our benchmark provides an open, stack-agnostic testbed for evaluating arbitrary frontier coding agents on autonomous 3D game production.

\section{Task Specification Format}
\label{app:task-specification}

Every task specification follows the five-part schema introduced in Section~\ref{sec:task-specifications}. Table~\ref{tab:task-specification} shows how the five parts are instantiated for two representative \emph{Hard}-tier 3D vertical slices.
\subsection{Acceptance Criteria}
\label{app:acceptance-criteria}
The Acceptance Criteria section is written after the first four sections of a specification are final.
A separate agent takes these four sections as input and writes the fifth: clear, reproducible, objectively verifiable test cases in natural language. No game exists at this point. The prompt below also fixes the rules that test code must follow later, when the criteria are run against a finished game, which is why it refers to game source code. In short, tests are non-visual: they simulate real inputs in a headless browser and read runtime state and logs to check numerical rules, map collisions, input responses, and state transitions. They are derived from the requirements, never from a game's own code, and they run on a copy of the delivered game, whose functions and rules they must not change.

\begin{promptbox}{Non-Visual Basic Functional Acceptance Constraints Prompt}

You are responsible for formulating multiple basic functional acceptance criteria based on the given game Prompt. Focus on determining whether core operations, numerical rules, maps and collisions, and state transitions meet the requirements. Output clear, reproducible, and objectively verifiable natural-language test case descriptions, without generating game implementation code.

\medskip
\noindent\textbf{Game Code and Browser Constraints}
\begin{itemize}
    \item Game source code, configurations, maps, assets, and dependency files must all remain read-only. Subsequent test implementations may only write or modify test-side code, configurations, and reports.
    \item Adding game debug interfaces is permitted, but you must strictly refrain from modifying existing game functions, replacing game functions, altering collision or win/loss rules, or directly setting states such as health, gold, position, or cooldowns to establish test scenarios. Preconditions must be reached through existing game operations.
    \item Invoking interactive Chrome Debug / Chrome DevTools tools, visible browser windows, or existing user browser sessions is prohibited. Test scripts are permitted to launch an isolated \textbf{headless Chromium/Chrome} instance on their own to conduct programmatic testing and read-only inspection via Playwright or that instance's CDP.
    \item Using Playwright's \texttt{--headed}, \texttt{--ui}, \texttt{--debug}, and Inspector is prohibited; among these, \texttt{--debug} opens a browser window and Inspector (\href{https://playwright.dev/docs/running-tests}{Playwright Official Documentation}).
    \item Do not use screenshots, videos, pixels, colors, image recognition, or manual visual inspection as acceptance evidence. Real keyboard and mouse inputs, DOM/accessibility text, existing runtime states, events and console logs, as well as read-only inspection of source code and static data, are permitted.
\end{itemize}

\medskip
\noindent\textbf{Basic Functional Testing Focus}
\begin{itemize}
    \item \textbf{Values and Resources}: Check the initial values, increments/decrements, consumption, caps, damage, rewards, cooldowns, and settlement rules declared in the source Prompt.
    \item For relevant operations, separately check insufficient resources, exact resources, surplus resources, as well as repeated operations, invalid targets, and operations during cooldowns.
    \item Check whether a single valid event produces the specified number of deductions, damage, or rewards, and whether failed or rejected operations maintain the states that should remain unchanged.
    \item Expected values must be derived independently from the Prompt; do not copy calculation results from the game implementation as the correct answer. For values, units, boundaries, or tolerances not specified in the source text, do not invent them yourself. Numerical test cases shall use equivalence partitioning and boundary value analysis (\href{https://www.istqb.org/wp-content/uploads/2024/11/ISTQB_CTFL_Syllabus_v4.0.1.pdf}{ISTQB Testing Techniques, Section 4.2}).
    \item \textbf{Maps and Collisions}: Check the scenes, spawn points, objectives, obstacles, zones, and map transitions required by the source Prompt.
    \item Verify whether spawning and movement are valid, whether collision and out-of-bounds handling are correct, and whether mandatory objectives can be reached through normal operations.
    \item For readable map data, graph search may be used to check paths; conditions such as passage direction, character colliders, jumping ability, doors, and keys must be considered---do not treat all maps as undirected planar grids. Separately record ``a path exists in the map data'' and ``the character can traverse the path via real operations.'' The existence of configurations, navigation returning a path, or the presence of scene objects cannot individually prove that runtime behavior is correct.
    \item \textbf{Operations and States}: Verify through real inputs that primary operations actually change the corresponding states; do not merely check that functions or event listeners exist.
    \item Cover start, pause, resume, win/loss, failure, retry, and scene transitions according to the source Prompt, checking for illegal phase operations, duplicate submissions, and residual old inputs.
    \item Include at least one reachable core gameplay loop. Fixed seeds and deterministic repetition are checked only if already supported by the game or explicitly required by the Prompt; do not override random numbers or the game clock to manufacture consistent results.
\end{itemize}

\medskip
\noindent\textbf{Test Case Writing and Verdict Criteria}
\begin{itemize}
    \item Each test case verifies only one independent objective and includes: ID, source clause reference, preconditions, actions, evidence, pass criteria, and fail criteria.
    \item Provide positive, rejection, and boundary scenarios for key rules; isolate other mechanisms to prevent them from masking the result under test.
    \item Do not use undefined vague criteria such as ``normal,'' ``nearby,'' ``sufficient,'' or ``obvious.'' When rules conflict or lack basis, list only the affected sub-assertions as to-be-confirmed.
    \item Static evidence can only support static conclusions; dynamic features must have evidence of actual operations and state changes. The test model should be a simplified model built from requirements, not a direct copy of the implementation under test. When existing states or logs cannot be obtained, clearly state the evidence gap and mark it as \texttt{BLOCKED}; do not modify the game to adapt to the test, nor directly mark it as \texttt{PASS} or \texttt{FAIL}.
    \item Mark unexecuted test cases as pending execution; do not claim they have already passed. After execution, record actual inputs, pre- and post-states, expected values, actual values, and failure reproduction steps.
\end{itemize}

\end{promptbox}
\clearpage

\subsection{Sources and reference images.}
Tasks are curated from widely played, top-charting commercial titles across major distribution ecosystems, including Steam, online PC platforms, and console storefronts. Where a task provides reference images, they communicate intended scene composition, visual aesthetics, or spatial layouts. They serve strictly as non-executable design constraints: a reference image neither supplies an interactive scene graph nor establishes that a generated build satisfies functional requirements.

\begin{table*}[tp]
\centering
\caption{Canonical five-part structure of benchmark task specifications, illustrated with two representative \emph{Hard}-tier 3D vertical slices: a MOBA lane duel (\emph{1v1 Nexus Duel}, modeled after \emph{League of Legends}) and an arcade kart racer (\emph{Town Highway Circuit}, modeled after \emph{Crazyracing KartRider}).}
\label{tab:task-specification}
\footnotesize
\setlength{\tabcolsep}{5pt}
\renewcommand{\arraystretch}{1.22}
\begin{tabularx}{\textwidth}{
  >{\raggedright\arraybackslash\bfseries}p{0.15\textwidth}
  >{\raggedright\arraybackslash}p{0.21\textwidth}
  >{\raggedright\arraybackslash}X
  >{\raggedright\arraybackslash}X
}
\toprule
\textbf{Specification Section} & \textbf{Formal Purpose} & \textbf{Instantiation A: MOBA Slice (\emph{1v1 Nexus Duel})} & \textbf{Instantiation B: Arcade Racing Slice (\emph{Town Highway})} \\
\midrule
Role and Objective &
Defines the target software artifact, player/AI configuration, and functional task scope. &
Construct a playable 3D browser MOBA duel pitting one human-controlled champion against an autonomous AI opponent, featuring six selectable champions and terminating upon Nexus destruction. &
Construct a single-player 3D arcade kart-racing adaptation featuring the \emph{Town Highway} circuit, three selectable drivers, a \emph{Cotton PRO} kart, and a two-lap race against three AI competitors. \\
\addlinespace[3pt]
Game Overview and Player Experience &
Articulates the intended player experience, core loop pacing, mechanical highlights, and visual direction. &
Deliver an 8--15-minute tactical lane duel centered on minion last-hitting, skillshot evasion, damage trading, itemization, and wave-assisted structure pushing. &
Deliver a 2--3-minute race set in a vibrant cartoon town, emphasizing corner entry, drift initiation, counter-steering, grip recovery, and strategic nitro-booster overtaking. \\
\addlinespace[3pt]
Project Setup and Deliverables &
Establishes engine/rendering constraints, build contracts, and self-contained packaging rules. &
Deliver complete source code and a root-level \texttt{build.sh} script compiling an entry \texttt{index.html}, with all runtime assets bundled locally offline. &
Deliver complete source code and a root-level \texttt{build.sh} compiling \texttt{index.html} with all runtime assets bundled locally. \\
\addlinespace[3pt]
Complete Game Design &
Formalizes controls, state variables, numerical parameters, state transitions, and termination rules. &
The Nexus holds $2{,}500\text{ HP}$ and remains invulnerable while its defensive lane turret stands. Champion eliminations create wave-pushing windows rather than immediate victory. &
Sustained moving drifts charge a $[0, 100]$ gauge; reaching $100$ grants one booster (up to two stored slots). Pressing \texttt{Ctrl} consumes one booster only when no boost is currently active. \\
\addlinespace[3pt]
Acceptance Criteria &
Translates design rules into deterministic, reproducible checks with observable runtime outcomes. &
Attacking the Nexus while its turret is intact must leave Nexus HP unchanged; destroying the turret and subsequently reducing Nexus HP to zero must trigger the victory state while logging all HP deltas and state transitions. &
Verify that stationary steering yields zero gauge charge. Execute valid drifts to earn and trigger a booster: stored count must decrement by one and kart velocity must increase, while repeated \texttt{Ctrl} inputs during active boost must neither consume extra stock nor extend boost duration. \\
\bottomrule
\end{tabularx}
\end{table*}

\section{Functional Evaluation: Details}
\label{app:functional-details}

This appendix gives the details of the three stages of Section~\ref{sec:code-acceptance}; Figure~\ref{fig:acceptance-harness} shows a worked example.


\paragraph{Illustrative content of Figure~\ref{fig:framework}.}
The Borderlands note, the harness--model pairs, and the MOBA subitem in Figure~\ref{fig:framework} are illustrative: the two candidate rows of the subitem are historical assertions and the third is synthetic, not a current-protocol result.

\paragraph{Inputs and scope.}
The protocol takes the acceptance criteria, which a separate agent derives from the finished specification before any candidate is built (Appendix~\ref{app:acceptance-criteria}), and the registered source code of the candidate. Registration records provenance; it does not assert byte-identity with a published build. Tests run on a copy of the delivered game, whose source stays read-only. The channel uses no visible browser window, pixels, screenshots, visual classifiers, or model judges: it calls production handlers and code-level view models directly or drives them with scripted inputs in a headless browser, and may read DOM text, runtime state, events, and logs. It therefore verifies state logic, not screen visibility or usability.

\paragraph{Writing and auditing tests (Stage 2).}
Given the frozen subitem IDs, the candidate's codebase is inspected (interfaces, state, clocks, and entry points), and multiple independent, randomized tasks per subitem write unit tests adapted to that codebase. The reviewer agent that audits the tests revises them at most once and modifies only test or driver code. Programmatic checks on subitem IDs, allowed files, and syntax run alongside the review. The result is multiple assertions per subitem.

\paragraph{Execution (Stage 2).}
An isolated worker executes the assertions one by one against real production functions, commands, or event paths on fresh fixtures; an engine-native driver may expose state without replacing the tested logic. The worker logs inputs, before and after states, events, resource changes, observed and expected values, tolerances, and clock boundaries. Expected values must come independently from the specification and legal preconditions, never from observed outputs or the candidate's tested formula. Source-text checks apply only to inherently static rules, and passing requires actual assertion execution, not author claims or forced outcomes. Real functional failures remain failures, and infrastructure errors are logged separately.

\paragraph{Validity and scoring (Stage 3).}
The set $V_g$ of validly evaluated criteria excludes explicit, spec-grounded \texttt{OUT\_OF\_SCOPE} items, \texttt{TEST\_ERROR}s, unexecuted \texttt{BLOCKED} claims, infrastructure errors, and missing, stale, or unrun results. For $c\in V_g$, subitem $j\in\{1,\dots,m_c\}$ gets $b_{g,c,j}=1$ if and only if all of its assertions execute and pass, and $0$ otherwise, including when worker execution proves it \texttt{BLOCKED}; $k_{g,c}=\sum_{j=1}^{m_c}b_{g,c,j}$. Each passed subitem earns $1/m_c$ of its criterion (for example $1/7$ when $m_c=7$), regardless of how many assertions were generated for it. Criteria carry equal weight, and legacy weight fields are unused. Across a configuration's declared task set, the overall score is the equal mean of $\operatorname{FunctionalScore}(g)$ over games with defined scores (undefined if there are none), which keeps real zero scores in the mean and excludes invalid or missing runs.

\section{3D Presentation and Playability: Details and Validation}
\label{sec:player-experience-appendix}

\subsection{Presentation scoring}
\label{sec:presentation-scoring}
The reference set has four games from three tasks; the references themselves are not scored. Each game is compared with every reference twice, once in each left--right order, giving eight comparisons. A judged margin of slight, clear, or large counts as an outcome of $0.70$, $0.85$, or $0.95$ for the winner, and a tie as $0.5$. We estimate $\theta_g$ with a weak prior toward the scale center ($\lambda=0.2$) and give a score only when all comparisons are valid. A game that loses all eight comparisons by a clear margin scores $17.2$.

\subsection{Playability scale}
\label{sec:playability-rubric}
For each moment, the judge receives the definition of one aspect, the five levels in Table~\ref{tab:playability-rubric}, one short example per level, and the moment itself. The examples come from genres outside the benchmark, such as card games or golf, so that the judge compares experiences rather than mechanics. The judge first describes what the player sees, then picks a level and explains why it is not the neighboring one. 


\begin{table}[htbp]
\centering
\scriptsize
\setlength{\tabcolsep}{3pt}
\caption{Five-level playability scale (translated from the judge prompt).}
\label{tab:playability-rubric}
\begin{tabularx}{\linewidth}{@{}cYY@{}}
\toprule
Level & Control: does the action get a readable response? & Flow: does the player know the state and the next step? \\
\midrule
1 & The player probably cannot tell whether the action worked, or misreads the result. & The player probably misses the change, or thinks the game is stuck or broken. \\
2 & Something changes, but the player has to search or guess what it means. & The player notices a change but has to guess the state or the next step. \\
3 & The result is recognizable, but feedback is weak, far from where the player is looking, or hard to tell apart. & The state is clear, but the player has to explore to find out how to continue. \\
4 & The result is clear, timely, and where the player is looking; only minor roughness. & State and next step are both clear; only small details are missing. \\
5 & As 4, and the kind, size, and target of the result are obvious at a glance. & As 4, and the player also knows what to expect, such as a time limit or the cost of failing. \\
\bottomrule
\end{tabularx}
\end{table}

Table~\ref{tab:presentation_playability} gives the per-configuration scores behind Takeaways~8 and~9.
\begin{table}[htbp]
\centering
\caption{\textbf{Automated 3D Presentation and checkpoint Playability.} \textbf{3D Pres.} ($0$--$100$) measures reference-anchored visual scene quality (mean, with median in parentheses). \textbf{Playability} ($0$--$100$) averages \textbf{Control} and \textbf{Flow} ($1$--$5$ rubric means, rescaled to $0$--$100$ in parentheses).}
\label{tab:presentation_playability}
\resizebox{\linewidth}{!}{%
\begin{tabular}{lccccc}
\toprule
\textbf{Config.} & \textbf{3D Pres.} & \textbf{Playability} & \textbf{Control} & \textbf{Flow} & \textbf{BT} \\
\midrule
\texttt{gpt-6.1-sol} & 57.8 {\scriptsize(61.4)} & 68.4 {\scriptsize(68.8)} & 3.40 {\scriptsize(60.1)} & \textbf{4.07} {\scriptsize(\textbf{76.8})} & 69.11 \\
\texttt{gpt-6-astra} & \textbf{58.5} {\scriptsize(59.9)} & \textbf{69.5} {\scriptsize(\textbf{70.0})} & \textbf{3.49} {\scriptsize(\textbf{62.2})} & \textbf{4.07} {\scriptsize(76.7)} & \textbf{69.24} \\
\texttt{kimi-k3}     & 42.3 {\scriptsize(43.4)} & 64.9 {\scriptsize(65.0)} & 3.35 {\scriptsize(58.7)} & 3.84 {\scriptsize(71.1)} & 25.86 \\
\texttt{glm-5.3}     & 23.5 {\scriptsize(17.2)} & 54.9 {\scriptsize(55.8)} & 2.88 {\scriptsize(47.1)} & 3.51 {\scriptsize(62.6)} & 35.78 \\
\bottomrule
\end{tabular}%
}
\end{table}

\subsection{Configuration comparisons}
\label{sec:player-experience-pairs}
Table~\ref{tab:px-pairs} gives the paired differences behind the results. Each configuration covers a different subset of tasks, so we compare two configurations only on the tasks where both have both scores, and report the mean difference with a 95\% percentile bootstrap interval over tasks. Reference games are excluded.
\begin{table}[htbp]
\centering
\scriptsize
\setlength{\tabcolsep}{2.5pt}
\caption{Paired differences (first minus second configuration) on shared tasks. Brackets: 95\% bootstrap interval over tasks (2{,}000 draws).}
\label{tab:px-pairs}
\resizebox{\linewidth}{!}{%
\begin{tabular}{@{}lcc@{}}
\toprule
Pair & Pres.\ diff & Play.\ diff \\
\midrule
\texttt{gpt-6.1-sol} -- \texttt{gpt-6-astra} & $-2.0$ $[-5.3, +1.2]$ & $-1.4$ $[-3.1, +0.5]$ \\
\texttt{gpt-6.1-sol} -- \texttt{kimi-k3} & $+17.4$ $[+13.1, +21.8]$ & $+4.2$ $[+1.9, +6.6]$ \\
\texttt{gpt-6.1-sol} -- \texttt{glm-5.3} & $+32.4$ $[+26.7, +38.2]$ & $+10.6$ $[+7.0, +14.0]$ \\
\texttt{gpt-6-astra} -- \texttt{kimi-k3} & $+17.8$ $[+12.4, +23.2]$ & $+5.2$ $[+3.0, +7.3]$ \\
\texttt{gpt-6-astra} -- \texttt{glm-5.3} & $+34.8$ $[+29.0, +40.3]$ & $+13.5$ $[+9.9, +17.0]$ \\
\texttt{kimi-k3} -- \texttt{glm-5.3} & $+20.2$ $[+14.0, +26.4]$ & $+8.7$ $[+5.0, +12.6]$ \\
\bottomrule
\end{tabular}%
}
\end{table}

\subsection{How moments were reached}
\label{sec:player-experience-arrival}
Across the four configurations in the results, $37\%$ of the recorded moments were reached naturally and $63\%$ with assistance; about $6\%$ of the planned moments were not recorded. About half of the assisted moments only had their precondition set through the adapter, with no help inside the recorded window, so all feedback in the frames comes from the game itself; in the other half the adapter was also used during the recording. Per-game reports keep all three labels. The narrow range of playability across configurations is not an artifact of assisted arrival: recomputed from naturally reached moments alone, playability is $69.3$, $68.3$, $63.7$, and $54.5$ for \texttt{gpt-6-astra}, \texttt{gpt-6.1-sol}, \texttt{kimi-k3}, and \texttt{glm-5.3}, close to the values over all moments ($69.5$, $68.4$, $64.9$, $54.9$). Each moment is rated on an absolute five-level scale, and $76\%$--$85\%$ of the rated moments of every configuration fall at level $3$ or $4$.

\subsection{Worked example}
\label{sec:player-experience-example}
\begin{figure*}[htbp]
\centering
\input{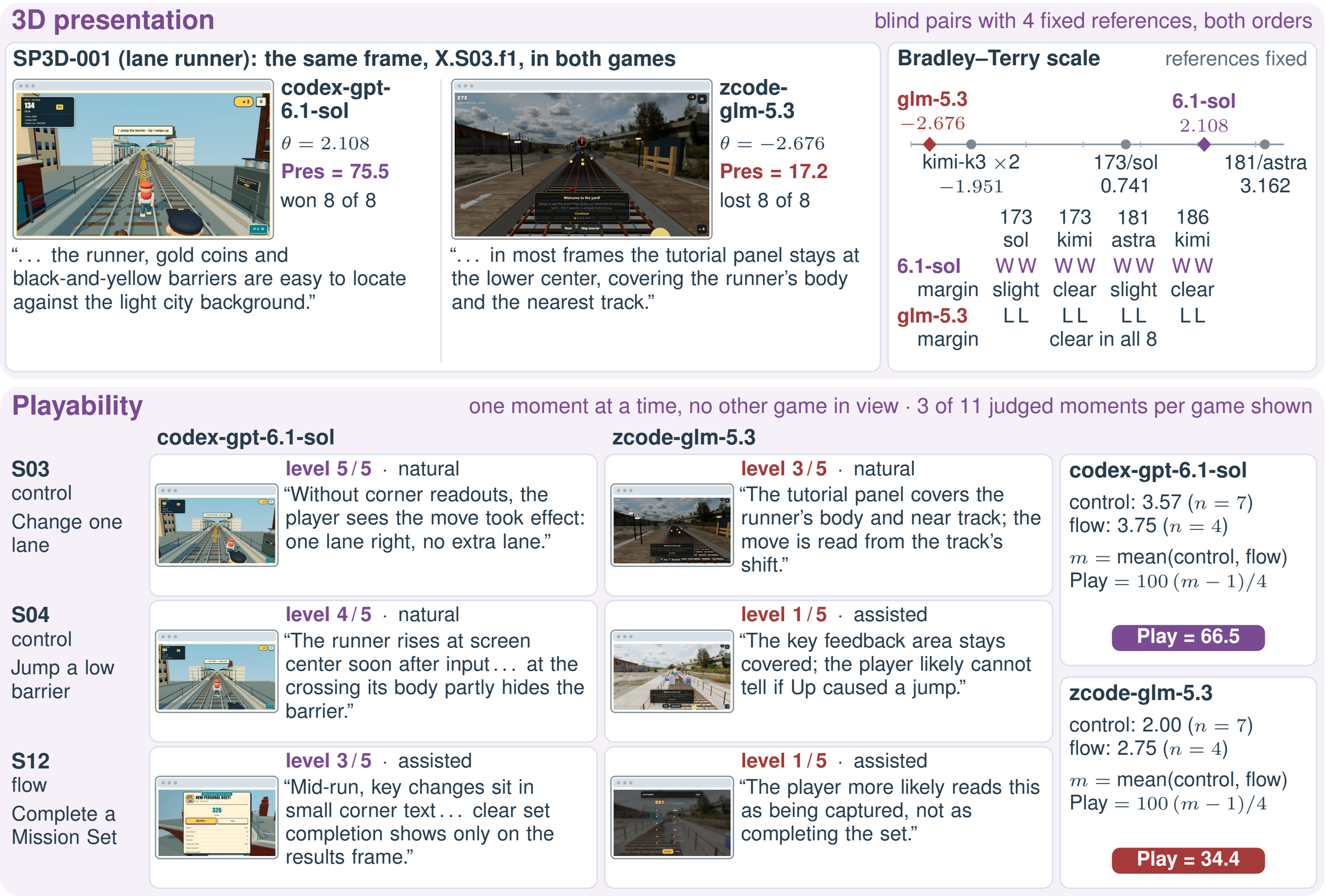}
\caption{3D presentation and playability evaluation. Key situations are planned from the task specification alone, then played and recorded once per game. Example: two agents on the same task, SP3D-001 (lane runner): codex-gpt-6.1-sol and zcode-glm-5.3. Top: the same moment in both games; the judge faults zcode-glm-5.3 for a tutorial panel that keeps covering the runner and the near track. On the Bradley--Terry scale the reference strengths are fixed. codex-gpt-6.1-sol ($\theta=2.108$) wins all 8 comparisons with the 4 references and scores 75.5 on presentation; zcode-glm-5.3 ($\theta=-2.676$) loses all 8 clearly and scores 17.2. Bottom: the same 3 of 11 judged moments in each game, with an excerpt of the judge's reason and how the moment was reached; playability is 66.5 versus 34.4. Quotes are translated from the judge's output and shortened.}
\label{fig:evaluator}
\end{figure*}

Figure~\ref{fig:evaluator} traces one task through situation planning, recording, and both judging formats.

\begin{table*}[tp]
\centering
\small
\setlength{\tabcolsep}{4pt}
\caption{Validation of the 3D presentation and playability scores on an early subset , run before the full evaluation. Agreement counts same-task game pairs whose order matches the direct pairwise comparison (agree\,/\,reverse\,/\,tie). The perturbation rows edit one game and replay the same situations.}
\label{tab:px-meta}
\begin{tabularx}{\linewidth}{@{}p{0.2\linewidth}Yp{0.3\linewidth}@{}}
\toprule
Question & Test & Result \\
\midrule
Is it stable? & Rate the same moments twice & identical in all cases \\
 & Swap the per-level examples for a different set, or drop them & game scores almost unchanged ($r\approx0.95$) \\
\addlinespace
Does each score track its own property? & Remove the game's hit and action feedback effects & control drops; flow and presentation unchanged \\
 & Repaint the scene structure in one flat color & presentation drops; control and flow unchanged \\
\addlinespace
Why two formats? & Rate presentation one moment at a time & order agreement drops (23\,/\,8\,/\,2); the repaint goes unnoticed \\
 & Judge playability by pairwise comparison & the repaint also lowers control and flow \\
\bottomrule
\end{tabularx}
\end{table*}

The last two rows explain the split in Section~\ref{sec:player-experience}. Rated one moment at a time, presentation misses an obvious repaint: the judge even mentions the flat colors in its description, yet gives the same level. Seen side by side, the change is easy to spot. Playability shows the opposite pattern. Under pairwise comparison, the repaint leaks into control and flow, while per-moment rating ignores it and still picks up the removed feedback.
\section{Human Preference: Details}
\label{app:preference-details}

This appendix completes Section~\ref{sec:preference}. Responses are a preference for either game, a tie, both games bad, or skip. Ties and both-bad responses remain distinct in the records but each contributes a half-win to both configurations; skips contribute no evidence to the ranking.

\paragraph{Fitting the strengths.}
The BT Score of Equation~\ref{eq:human-preference} is the expected win rate of configuration $a$ against a uniform average over the other $K-1$ configurations ($K>1$). The parameters are fitted by maximizing a weighted, L2-regularized log-likelihood over the valid votes,
\begin{equation}
\begin{aligned}
\hat{\beta}=\arg\max_{\beta}\;&\sum_{v} w_v\Big[y_v\log\sigma(\beta_{a_v}-\beta_{b_v})\\
&+(1-y_v)\log\sigma(\beta_{b_v}-\beta_{a_v})\Big]-\frac{\lambda}{2}\,\lVert\beta\rVert_2^2,
\end{aligned}
\label{eq:bt-fit}
\end{equation}
where $v$ indexes the valid votes, $a_v$ and $b_v$ are the two configurations shown in vote $v$, and $y_v\in\{1,\tfrac12,0\}$ is the outcome for $a_v$: $1$ if $a_v$ wins, $0$ if $b_v$ wins, and $\tfrac12$ for a tie or both-bad. Skipped votes are not valid votes and do not enter the sum.

\paragraph{Sampling weights and regularization.}
The weight $w_v$ corrects for matchups being drawn with unequal probability. Let $p_v$ be the conditional probability that the arena's sampler selected the type of matchup that vote $v$ belongs to. We set
\begin{equation}
\tilde{w}_v=\frac{1}{p_v},\qquad w_v=\frac{N\,\tilde{w}_v}{\sum_{u}\tilde{w}_u},
\label{eq:bt-weights}
\end{equation}
where $u$ runs over the $N$ valid votes, so that the weights have mean one. A rarely sampled matchup thus counts more per vote. The weight depends only on the sampling design: it is neither a measure of rater reliability nor a model score. The L2 term keeps the strengths finite when a configuration wins or loses every comparison it appears in. Its strength is fixed at $\lambda=10^{-3}$ and is not learned from the votes: a larger $\lambda$ shrinks the strengths toward zero and gives more conservative scores, whereas a smaller $\lambda$ follows the observed votes more closely but becomes more extreme with few votes.

\paragraph{Raw Win.}
\emph{Raw Win} is the unweighted share of a configuration's valid comparisons that it won, with ties and both-bad responses counted as half-wins for both sides:
\begin{equation}
\operatorname{RawWin}(a)=\frac{W_a+\tfrac12 T_a+\tfrac12 B_a}{N_a},
\label{eq:raw-win}
\end{equation}
where $W_a$, $T_a$, and $B_a$ are the numbers of clear wins, ties, and both-bad responses of configuration $a$, and $N_a$ is the number of all its non-skipped comparisons; skips are removed from both the numerator and the denominator. Raw Win uses neither the sampling weights $w_v$ nor any per-voter weighting, and it does not use the strength model, so unlike the BT Score it does not account for which opponents a configuration happened to meet.

\paragraph{Reporting.}
Preference measures relative player judgment and is reported separately from both automatic scores and generation success. A completed study must disclose sampling probabilities, repeat-vote handling, valid vote counts, and uncertainty clustered by participant or anonymous session.

\section{Skill-Augmented Generation Protocol and Prompts}
\label{app:generation_protocol}

This appendix details the modular skills, 3D asset acquisition rules, and exact orchestration prompts used in the two-stage game generation protocol introduced in Section~\ref{sec:protocol}. Agent Skills~\citep{wang2024voyager, anthropic2025agentskills, xu2026agentskills, li2026skillsbench} are composable packages of procedural workflows, verification scripts, and domain constraints that an agent loads progressively on demand.

\begin{table}[htbp]
\centering
\footnotesize
\setlength{\tabcolsep}{3pt}
\caption{Evaluated configurations. Short names are used in the text and in Tables~\ref{tab:token_decomp}--\ref{tab:asset_strategy}.}
\label{tab:configs}
\resizebox{\linewidth}{!}{%
\begin{tabular}{@{}llll@{}}
\toprule
Configuration ID & Short name & Harness & Model \\
\midrule
\texttt{codex-gpt-6.1-sol} & \texttt{gpt-6.1-sol} & Codex CLI 0.167.0 & GPT-6.1 Sol \\
\texttt{codex-gpt-6-astra} & \texttt{gpt-6-astra} & Codex CLI 0.167.0 & GPT-6 Astra \\
\texttt{zcode-glm-5.3} & \texttt{glm-5.3} & Zcode 0.16.9 & GLM-5.3 \\
\texttt{kimi-k3} & \texttt{kimi-k3} & Kimi CLI 2.6.1 & Kimi K3 \\
\bottomrule
\end{tabular}%
}
\end{table}

\paragraph{Evaluated configurations.}
Table~\ref{tab:configs} lists the four evaluated configurations with their harness versions. Each pairs a native coding-agent harness with one model and runs at the highest reasoning setting available in its harness. Each model runs in exactly one harness, so a short name identifies a configuration.

\subsection{Two-Stage Skill Workflow and Asset Pipeline}
\label{app:skill-workflow}

\paragraph{Stage 1: Architectural Planning and Asset Orchestration.}
Given a benchmark task specification, a lead orchestration agent inspects the target repository structure, manifest schema, and build contract without writing production game code. Domain-level coordination is governed by the  iterative browser verification guidelines from \texttt{develop-web-game}~\citep{openai2026developwebgame}. For 3D tasks, dedicated subagents are spawned to prepare environmental meshes and textures. Subagents prioritize retrieving open, royalty-free runtime assets formatted in \textbf{glTF~2.0} ~\citep{khronos2022gltf}—utilizing compact binary (\texttt{.glb}) scene hierarchies, Physically Based Rendering (PBR) materials. Only when suitable public structural models cannot be located do subagents invoke \texttt{meshy-3d-generation}~\citep{meshy2026agent} for text- or image-to-3D synthesis (capped at five generated models per project). Once technical risks and workspace identifiers are resolved, the lead agent invokes \texttt{writing-plans}~\citep{vincent2025superpowers} to produce a concrete, file-level implementation blueprint coupling a single unified simulation state to controls, physics, onboarding UI, and acceptance checks.

\paragraph{Stage 2: Subagent-Driven Plan Execution.}
An execution agent receives the approved Stage~1 blueprint and produces the final deliverable using \texttt{superpowers:subagent-driven-development} and \texttt{superpowers:executing-plans}~\citep{vincent2025superpowers}. To prevent context degradation across multi-file implementations, \texttt{subagent-driven-development} dispatches isolated implementer subagents per planned task, enforces post-task specification-compliance and code-quality review gates, and records state transitions in a persistent progress ledger. Throughout execution, agents perform live headless-browser debugging and interactive playtesting~\citep{openai2026developwebgame}—prioritizing simulation fidelity and gameplay correctness over raw frame rate ($\ge 20\text{ FPS}$)—until the root \texttt{build.sh} script compiles a self-contained \texttt{index.html} satisfying all acceptance criteria.
\clearpage
\onecolumn
\subsection{Orchestration and Execution Prompts}
\label{app:prompts}

\begin{promptbox}{Stage 1: Lead Orchestration and Planning Prompt}
You are the lead orchestration agent responsible for generating this browser game. Use \texttt{\$threejs-game-director} as the primary orchestration skill, alongside \texttt{\$develop-web-game} for iterative browser verification, \texttt{\$writing-plans} for architectural planning, \texttt{\$threejs-3d-generator} for structural 3D models, and automatically invoke relevant gameplay, graphics, UI, asset, audio, browser-debugging, performance, and QA skills when appropriate.

\vspace{3pt}
\textbf{Hard Constraints:}
\begin{itemize}[leftmargin=1.3em, itemsep=2pt, topsep=3pt, parsep=0pt]
    \item \textbf{Session Isolation:} Conduct all generation within this conversation thread; do not invoke external platform-level automated generation workflows.
    \item \textbf{Contract Discovery:} Inspect the repository first to infer the exact artifact format, directory layout, manifest schema, build contract, and database linkage expected by the harness (e.g., \texttt{\textless WORKSPACE\_ROOT\textgreater/games/\textless TASK\_ID\textgreater/\textless HARNESS\_MODEL\textgreater/\textless RESULT\_ID\textgreater/}).
    \item \textbf{Database Integrity \& Zero Cross-Contamination:} Do not alter benchmark task prompts or generator scripts, confirm \texttt{\textless TASK\_ID\textgreater} before updating records, and never read or reuse files or assets from other games on the host machine.
    \item \textbf{Subagent Asset Pipeline \& glTF Delivery:} Spawn subagents to prepare environmental and texture assets, prioritizing open, royalty-free 3D assets in Khronos glTF~2.0 / \texttt{.glb} format (\url{https://www.khronos.org/gltf/}). Invoke \texttt{@meshy-3d-generation} only when suitable public structural meshes cannot be located (capped at $5$ generated models per project).
    \item \textbf{Onboarding \& QA Pass:} Include an interactive tutorial/onboarding overlay explaining objectives and controls, followed by a comprehensive code review, browser QA pass, and bug-fix iteration prior to packaging.
\end{itemize}

\textbf{Current Task (Planning Phase Only):} Do \emph{not} begin implementation yet. First, (1) determine the storage layout, manifest schema, build contract, and status-update mechanism; (2) audit available agent skills and local CLI tools; (3) identify technical bottlenecks, ambiguities, or contradictions in the specification; 

\textbf{Game Specification Input:} \texttt{\textless INSERT\_FIVE\_PART\_TASK\_SPECIFICATION\textgreater}
\end{promptbox}

\begin{promptbox}{Stage 2: Subagent-Driven Execution Handoff Prompt}

Implement the approved plan using \texttt{superpowers:subagent-driven-development} and \texttt{superpowers:executing-plans}
\end{promptbox}

\clearpage
\raggedbottom
\section{Long-Horizon Analysis: Protocol and Full Results}
\label{sec:retrospective-appendix}

This appendix explains how the goal retention (GR) and plan adherence (PA) numbers of Section~\ref{sec:long_horizon} were obtained and gives the full results.

\paragraph{What we sample.}
We work from the development logs of finished builds. An \emph{operation} is one step in which the main agent does real work with a tool, such as editing a file or running a command; steps that only wait, coordinate, or plan are skipped. We line up the operations of a run in order, cut them into four quarters by count (not by clock time), and draw up to eight at random from each quarter, which samples about $8\%$ of all operations. We also draw a smaller sample of operations by sub-agents; these are judged against the task the sub-agent was given and are kept out of the rates reported here. Two limits remain: the filter that selects operations misses some terminal commands issued through a wrapper, and the reviewer does not always see the full context around an operation.

\paragraph{How an operation is judged.}
A reviewer answers the two questions of Section~\ref{sec:long_horizon} for every sampled operation. Each answer is one of three labels: \emph{aligned} (A), the operation serves the goal or follows the plan; \emph{deviating} (D), there is clear evidence that it does not; or \emph{unclear} (U), the logs are not enough to tell. If the task is changed along the way, the goal changes with it, but the plan stays the first one the agent wrote. Operations carried out before that plan existed are left out of plan adherence, and an operation whose link to the plan cannot be established is unclear, not aligned.

\paragraph{Who reviews.}
The reviewers are LLM agents (Codex), not people. A random fifth of the sampled operations was reviewed a second time by a reviewer that did not see the first labels, and $99\%$ of these pairs are usable. On goal labels for main-agent operations the two reviews agree in $92.3\%$ of cases ($\kappa=0.443$; $\kappa$ is modest because almost all labels are A). On plan labels for sub-agent operations agreement is weak ($\kappa=0.270$), so sub-agent labels are not used for the rates in this paper. Every operation first labeled as deviating was also checked by a different reviewer; these extra checks are not part of the agreement figures. The two reviews are kept apart, but they are not fully blind to which configuration produced a run, and two LLM reviewers can make the same mistake. The tables use the labels after reconciliation.

\paragraph{From labels to rates.}
Because we draw at most eight operations per quarter, a sampled operation from a long quarter stands for more operations than one from a short quarter, and we weight it accordingly. With $A$ and $D$ the weighted counts of aligned and deviating operations in one build, we report
\[
\text{rate}=\frac{A}{A+D},
\]
the share of aligned operations among those that could be judged; this is the GR and PA of the main text. Unclear operations are left out of the rate. A configuration's value is the plain average over its builds, so every build counts the same however long it ran. A build whose logs cannot be reliably matched to it gives no rate. A rate of $100\%$ means that every sampled operation was aligned, not that every operation was.

\paragraph{Large tasks.}
Tasks are ranked by the structure of their specification, namely how many groups of requirements it contains and how long its longest chain of dependent steps is, and split into thirds: small, medium, and large. This reflects the size of the task, not how hard people find it or how long an agent takes.

\paragraph{Process rules.}
Goal retention also covers instructions about how to work, such as staying read-only, only planning during the planning stage, and where to put temporary files. Breaking such a rule counts as a deviation even when the agent is still building the right game. Table~\ref{tab:retrospective-sensitivity} therefore repeats the full-run goal retention with these violations counted as unclear instead of deviating (GR$^{\dagger}$). This does not excuse the violations, and the deviations that remain are not a pure measure of memory; it shows how much of a lower goal retention comes from process rules and not from the game itself.

\begin{table}[htbp]
\centering\small
\setlength{\tabcolsep}{4pt}
\caption{\textbf{Full-run goal retention} as judged (GR) and with process-rule violations set aside (GR$^{\dagger}$).}
\label{tab:retrospective-sensitivity}
\begin{tabular}{lrrrr}
\toprule
 & \multicolumn{2}{c}{All builds} & \multicolumn{2}{c}{Large tasks} \\ \cmidrule(lr){2-3}\cmidrule(lr){4-5}
Configuration & GR & GR$^{\dagger}$ & GR & GR$^{\dagger}$ \\
\midrule
\texttt{gpt-6-astra} & 100.00 & 100.00 & 100.00 & 100.00 \\
\texttt{gpt-6.1-sol} & 99.66 & 99.94 & 100.00 & 100.00 \\
\texttt{kimi-k3} & 96.06 & 99.28 & 94.51 & 97.64 \\
\texttt{glm-5.3} & 100.00 & 100.00 & 100.00 & 100.00 \\
\bottomrule
\end{tabular}
\par\smallskip{\footnotesize Values are percentages. For GR$^{\dagger}$, operations that broke a process rule are counted as unclear, not as aligned. }
\end{table}

\end{document}